\documentclass[lettersize,journal]{IEEEtran}
\usepackage{amsmath,amsfonts}
\usepackage{algorithmic}
\usepackage{algorithm}
\usepackage{array}
\usepackage[caption=false,font=normalsize,labelfont=sf,textfont=sf]{subfig}
\usepackage{textcomp}
\usepackage{stfloats}
\usepackage{url}
\usepackage{verbatim}
\usepackage{graphicx}
\usepackage{cite}
\usepackage{bbding}
\usepackage[colorlinks,linkcolor=blue]{hyperref}
\usepackage{amsmath}
\usepackage{multirow}
\usepackage{booktabs}

\begin{document}

\title{A Visual Representation-guided Framework with Global Affinity for Weakly Supervised Salient Object Detection}

\author{\IEEEauthorblockN{Binwei~Xu,
		%Haoran~Liang\mbox{*},
		Haoran~Liang,
		Weihua~Gong,
		Ronghua~Liang,~\IEEEmembership{Senior Member,~IEEE,}
		and Peng~Chen,~\IEEEmembership{Member,~IEEE}
		%and~Xiaofei~He,~\IEEEmembership{Senior Member,~IEEE}
	}\\
	% <-this % stops a space
	
	\thanks{Manuscript received 23 February 2023; revised 22 April 2023 and 17 May 2023; accepted 27 May 2023. This work was supported by the National Key Research and Development Program of China (2020YFB1707700, 2022YFB3304100), the National Natural Science Foundation of China (62176235, 62036009, 61871350, U1909203), and Zhejiang Provincial Natural Science Foundation of China (LY21F020026).
		
		Binwei Xu, Haoran Liang, Weihua Gong, Ronghua Liang, and Peng Chen are with the College of Computer Science and Technology, Zhejiang University of Technology, Hangzhou 310023, China (e-mail: \{xubinwei, haoran, whgong, rhliang, chenpeng\}@zjut.edu.cn). \textit{Corresponding author: Haoran Liang.}}% <-this % stops a space
	% (e-mail: \{xubinwei, haoran, rhliang, chenpeng\}@zjut.edu.cn)
	%\thanks{Xiaofei He is with Zhejiang University, Hangzhou 310023, China (e-mail: xiaofeihe@cad.zju.edu.cn).}%
	
	%\thanks{Corresponding author\mbox{*}: Haoran liang (email: haoran@zjut.edu.cn)}
}

% The paper headers
%\markboth{Journal of \LaTeX\ Class Files,~Vol.~14, No.~8, August~2021}%
%{Shell \MakeLowercase{\textit{et al.}}: A Sample Article Using IEEEtran.cls for IEEE Journals}

\IEEEpubid{0000--0000/00\$00.00~\copyright~2021 IEEE}
%\IEEEpubid{Copyright © 20xx IEEE. Personal use of this material is permitted. \\
%	\\
%	However, permission to use this material for any other purposes must be obtained from the IEEE by sending an email to
%	pubs-permissions@ieee.org.}
% Remember, if you use this you must call \IEEEpubidadjcol in the second
% column for its text to clear the IEEEpubid mark.

\maketitle

\begin{abstract}
Fully supervised salient object detection (SOD) methods have made considerable progress in performance, yet these models rely heavily on expensive pixel-wise labels. Recently, to achieve a trade-off between labeling burden and performance, scribble-based SOD methods have attracted increasing attention. Previous scribble-based models directly implement the SOD task only based on SOD training data with limited information, it is extremely difficult for them to understand the image and further achieve a superior SOD task. In this paper, we propose a simple yet effective framework guided by general visual representations with rich contextual semantic knowledge for scribble-based SOD. These general visual representations are generated by self-supervised learning based on large-scale unlabeled datasets. Our framework consists of a task-related encoder, a general visual module, and an information integration module to efficiently combine the general visual representations with task-related features to perform the SOD task based on understanding the contextual connections of images. Meanwhile, we propose a novel global semantic affinity loss to guide the model to perceive the global structure of the salient objects. Experimental results on five public benchmark datasets demonstrate that our method, which only utilizes scribble annotations without introducing any extra label, outperforms the state-of-the-art weakly supervised SOD methods. Specifically, it outperforms the previous best scribble-based method on all datasets with an average gain of 5.5\% for max f-measure, 5.8\% for mean f-measure, 24\% for MAE, and 3.1\% for E-measure. Moreover, our method achieves comparable or even superior performance to the state-of-the-art fully supervised models.

%Specifically, it outperforms the previous best scribble-based method on all datasets with significant improvements in max f-measure (5.5\%), mean f-measure (5.8\%), MAE (24\%), and E-measure (3.1\%). 

%Experimental results on five public benchmark datasets demonstrate that our method that only utilizes scribble annotations without introducing any extra label outperforms the state-of-the-art weakly supervised SOD methods , and
% is comparable or even superior to the state-of-the-art fully supervised models.
% outperforms previous best scribble-based method (SCWSSOD) on all datasets by a large margin with an average gain of 5.5\% for $F_{max}$, an average gain of 5.8\% for $F_{avg}$, an average gain of 24\% for $MAE$, and an average gain of 3.1\% for $E_m$
%Experimental results on five public benchmark datasets demonstrate that our method, which only utilizes scribble annotations without introducing any extra label,  outperforms the state-of-the-art weakly supervised SOD methods. In particular, it outperforms previous best scribble-based method on all datasets, achieving an average gain of 5.5\% in terms of $F_{max}$, an average gain of 5.8\% in terms of $F_{avg}$, an average gain of 24\% in terms of $MAE$, and an average gain of 3.1\% in terms of $E_m$. Furthermore, our method is comparable or even superior to the state-of-the-art fully supervised models. 

\end{abstract}


\section{Introduction}
\IEEEPARstart{S}{aliency} detection aims to identify the most visually distinctive objects from an image, which has rapidly developed and is widely applied in various vision fields such as image segmentation~\cite{ji2020encoder,wang2022looking}, object tracking~\cite{hong2015online}, image retrieval~\cite{30he2012mobile}, image editing~\cite{32cheng2010repfinder, 33cheng2017intelligent}, image cropping~\cite{wang2018deep}, and video segmentation~\cite{wang2017saliency}.
%Traditional works mainly achieve this task by utilizing hand-crafted feature, .
In recent years, fully supervised SOD methods~\cite{xu2021locate,8chen2018reverse,20pang2020multi,22zhang2017amulet,23qin2019basnet,24feng2019attentive} have made considerable progress in performance~\cite{wang2021salient}, but these models rely heavily on a large number of pixel-wise labels that are time-consuming and expensive to collect.
Therefore, to achieve a trade-off between labeling burden and performance, weakly supervised SOD methods have attracted increasing attention.

\begin{figure}[t]
	\centering
	\includegraphics[width=1\linewidth]{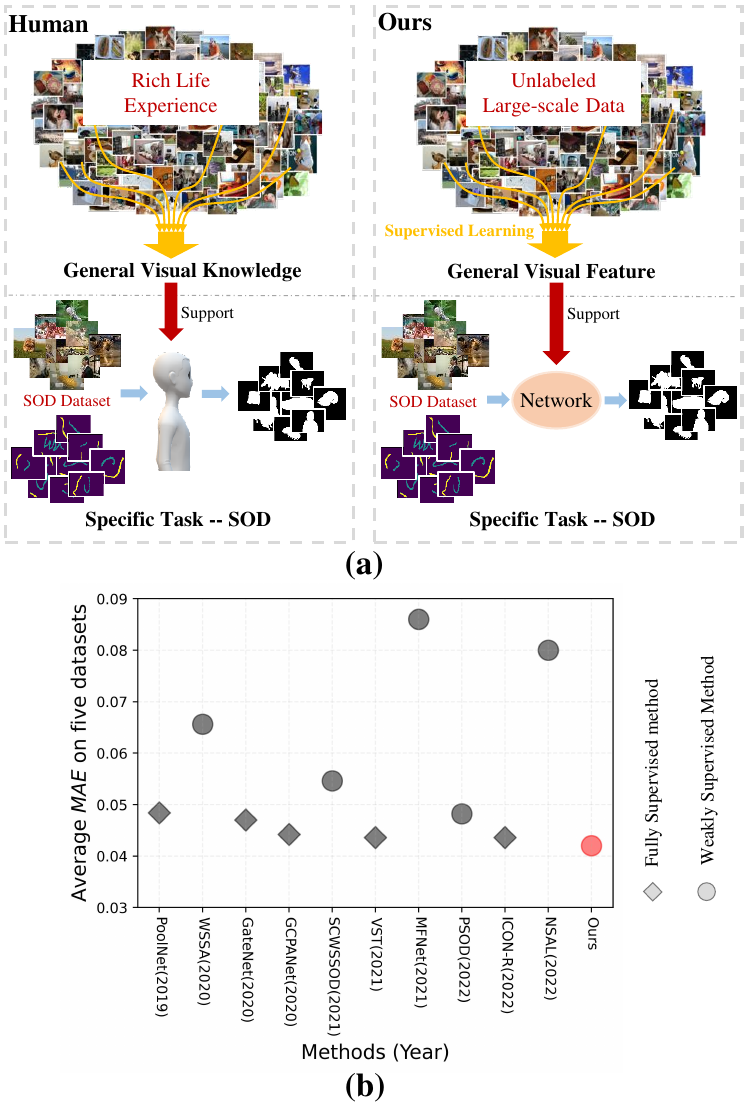} 
	\caption{(a) Illustration of our proposed framework guided by general visual features that simulate the general knowledge of humans. (b) Comparison of our method with the state-of-the-art fully and weakly supervised methods in terms of mean absolute error (MAE)~\cite{perazzi2012saliency} of five public SOD datasets. Our method lies at the bottom of the diagram and performs the best.
		%Sample results of our scribble-based SOD method are compared with SCWSSOD~\cite{yu2021structure}. 
		%Our method is trained on synthetic images and original images, while SCWSSOD is trained on original images. The results demonstrate that our method of introducing proposed synthetic images can perceive tortuous edges and predict a more accurate boundary.
		%By adding our proposed synthetic images as training data, our method can perceive tortuous edges and predict a more accurate boundary.
	}
	%	but SCWSSOD is trained on original images. Our model can perceive tortuous edges and predict more accurate boundary.}
\label{fig:coverfig}
\end{figure}

\IEEEpubidadjcol

Sparse labeling methods (e.g., image level, point, and scribble) have been proposed to relieve label burden while maintaining model performance. Image-level supervised SOD methods extract the class activation map (CAM)~\cite{zhou2016learning} from the image classification network as saliency localization. Nevertheless, the CAM contains rich noise that is almost impossible to correct in iterative training. Moreover, directly treating the most highlighted area in CAM as the salient region generated by the human eyes is somewhat controversial.
Point supervised labels~\cite{gao2022weakly} can provide local ground truth and are generated with lower labeling time cost. Although only a few points are annotated, these annotations are obtained based on the human vision system, which can correctly guide the model without introducing any noise. However, discerning the complete salient objects based on a few points, especially for the boundary of the salient objects, is difficult for the model. Therefore, \cite{gao2022weakly} uses an extra edge detector trained on edge datasets to help the model perceive object boundaries and predict integral salient objects. 
Moreover, scribble tags~\cite{zhang2020weakly} are a relatively cost-effective choice because they cover more regions of salient objects and background, and do not consume much time compared with point annotations. 
%\textbf{Accordingly, in this paper, we focus on scribble-based SOD method.}

Scribble annotations are located inside the salient objects and the background, so predicting the intact objects, especially the precise boundaries, remains a challenge.
\cite{zhang2020weakly} proposes the first scribble-based SOD method that applies an auxiliary edge detection task to help the model locate the object edges. However, this work introduces an additional edge detector and does not fundamentally address the issue that the scribble tag itself lacks edge information. 
Furthermore, SCWSSOD~\cite{yu2021structure} explores the intrinsic properties of images and presents a local saliency coherence (LSC) loss, which can extend the central scribble points to integral objects to some extent.
%and alleviate the problem of poor edge perception to some extent.
The core idea is that nearby pixels with similar color should exhibit similar saliency values.
%Although considerable progress has been achieved on scribble-based SOD, incomplete objects such as animal limbs and noisy boundaries still need to be improved. 
%Although considerable progress has been achieved on scribble-based SOD, incomplete objects predicted by models based on sparse labels remain much room for improvement. 
Although considerable progress has been achieved on scribble-based SOD, there is still much room for improvement in the incomplete objects predicted by such sparse label-based models due to the lack of global contextual perception.

However, humans given an image and the corresponding scribble tags can immediately and accurately distinguish which parts belong to the same object as the scribble label. The main difference between humans and the above models is general visual knowledge. Specifically, as shown in the Fig.~\ref{fig:coverfig}, humans possess rich life experience of various scenes and objects, so they can form a knowledge system, and it will support humans to construct the contextual connections between various regions of any image easily. 
%(\textbf{Humans will touch corresponding parts when they see one part})
%Humans are competent in such specific task based on the initial understanding of images, 
But previous models directly implement the SOD task only based on specific SOD training data without the support of rich experience and knowledge like humans, so breaking through the bottleneck is difficult for them.
%\textbf{Inspired by this biological capability, we first propose XXX that simulate the general visual cognition of humans for scribble-based SOD. (simple and effective)}
Therefore, we propose a simple yet effective framework guided by general visual representations that simulate the general visual knowledge of humans for scribble-based SOD.
%These general visual representations are generated by the vision transformer~\cite{dosovitskiy2020image} that is trained on large-scale unlabeled datasets in a self-supervised manner~\cite{caron2021emerging}, which is analogous to rich human experience, so they contain an initial understanding of images possess a wealth of contextual semantic information. 
These general visual representations are generated by the vision transformer (ViT)~\cite{dosovitskiy2020image} trained on large-scale unlabeled datasets in a self-supervised manner~\cite{caron2021emerging}, so they contain an initial understanding of images and possess a wealth of contextual semantic information. 
%For scribble-based SOD, understanding the image and further achieving a superior SOD task is extremely difficult for the model due to the limited information provided by the scribble tags.
Introducing general visual representations with rich semantic information can alleviate the above issues and support the model to understand and build connections between each part in an image.
%\textbf{Especially for our weakly supervised task, where the supervised information is already small, the model is hard to understand.}
%\textbf{task and understanding.}
% and understanding of the images.
In addition, some unsupervised SOD works~\cite{wang2022self, shin2022unsupervised} have begun to explore these features and validate their potential in SOD, but they directly use these features to determine the salient object rather than combine them with the network and lay the groundwork for understanding images.
%Additionaly, they still depend on hand-crafted priors to determine salient objects

%These SSL features are general image representations generated by learning on large-scale unlabeled datasets, which are like the rich experience of humans, so they contain rich contextual information about the relationships between each region of the image.
%\textbf{A, B, and C also have shown the success of the SSL feature.}
%\textbf{promising.}

%The introduction of general visual representation can effectively improve the discrimination of the model, but for the scribble label that covers only part of the object area, it is also crucial to develop a suitable loss function to guide the model to perceive the structure of the object.
Moreover, the scribble label covers only part of the object area, so developing a suitable loss function is also crucial to guiding the model to perceive the structure of the object. The previous study~\cite{yu2021structure} on scribble-based SOD only utilizes local information to expand the initial scribble regions via LSC loss, which penalizes pixels with similar color in local regions. Thus, propagating salient regions to complete salient objects in discontinuous or complex regions is difficult. In addition, based on the LSC loss and scribble annotation, which lack boundary-related supervision of salient objects, the model easily sacrifices indistinguishable concave regions and protruding elements, such as animal limbs, to guarantee correctness in predicting the majority of easily distinguishable regions of the salient objects. To address these issues, a global semantic affinity (GSA) loss, which further considers the overall affinity constraint between salient objects or backgrounds from a global perspective, is proposed. The main idea is to achieve high semantic similarities within the salient objects and within backgrounds by semantic features. In this way, regions far from the scribble or indistinguishable concave regions and protruding elements are directly linked to other regions and constrained by scribbles and other easy-to-identify areas during training.

In this paper, a visual representation guided-framework with global affinity, a simple but effective method for scribble-based SOD, is proposed. 
We employ ViT trained on large-scale unlabeled datasets by self-supervised learning to produce general visual representations. 
These representations, which are like rich human experience, contain abundant contextual semantic information and an initial understanding of images.
When we incorporate these features into the model, that is, based on a general understanding of the image, the model can better accomplish the weakly SOD task.
Unlike fully supervised methods, weakly supervised methods require a proper loss function to lead the model closer to the task requirements.
Here, we design a GSA loss to guide the model to aware contextual affinities from a global perspective; thus, it can capture more precise structure of salient objects.
Instead of low-level color features, we employ general visual representations to establish connections between image regions.

Despite its simplicity, our method substantially surpasses the said weakly supervised SOD methods. 
Moreover, our approach that only utilizes scribble annotations without introducing any extra label is comparable or even superior to the state-of-the-art fully supervised method. 
Fig.~\ref{fig:coverfig} shows the results of our method and the state-of-the-art fully and weakly supervised methods in terms of MAE of five public SOD datasets (DUTS-TE~\cite{wang2017learning}, PASCAL-S~\cite{li2014secrets}, ECSSD~\cite{2yan2013hierarchical}, HKU-IS~\cite{li2015visual}, and DUT-OMRON~\cite{yang2013saliency}) and our approach performs best.
Moreover, our method can be easily extended to point-based SOD, and the results prove the robustness and effective of our method.

Our main contributions are as follows:
\begin{itemize}
	\item [1.] 
	A novel scribble-based SOD architecture supported by general visual features is proposed, which simulates the rich life experience of the human, to help the model understand the contextual connections of images before implementing the SOD task.
	
	\item [2.] 
	A novel GSA loss function that maintains the high semantic similarities within the salient objects and within backgrounds is proposed, guiding the model to perceive the accurate structure of the salient objects.

	\item [3.] 
	The proposed method that only utilizes scribble annotations without introducing any extra label outperforms the state-of-the-art weakly supervised SOD methods. Moreover, It is comparable or even superior to the state-of-the-art fully supervised method.
	
\end{itemize}

The rest of the paper is structured as follows: Section~\ref{sec::related_work} provides a review of the related work on salient object detection, weakly supervised salient object detection, and object segmentation properties of self-supervised vision transformer. Section~\ref{sec::methodology} outlines the proposed methodology, Section~\ref{sec::experiment} presents the evaluation of the proposed method, Section~\ref{sec::limitation} demonstrates the limitation of our method, and Section~\ref{sec::conclusion} concludes the paper.
%通过全局约束，

%我们通过什么操作，解决了什么，具体点。

%The main reason is that the previous work does not further consider the overall affinity constraint between salient objects or backgrounds from a global perspective. 
%We find the main reason is that it does not further consider the overall affinity between salient objects or backgrounds from a global perspective. 
%Accordingly, we attack this issue 
%Based on this insight

%%%%%%%%%%%%%%%%%%%%%%%%%%
%Although the introduction of SSL features into unsupervised SOD \textbf{A, B, and C} has made some progress, the gap between these unsupervised methods and fully supervised SOD methods remains significant. 
%In particular, it is difficult for them to distinguish which objects are salient, especially in complex scenes. 
%The essence reason is that these unsupervised methods do not employ any strong manual annotation and these method based on hand-picked priors can't ensure the detected objects are salient.
%%therefore have to still rely on hand-picked priors with limited coverage such as \textbf{A, B, and C}.}
%\textbf{So ssl + scribble is promising, we settle the problem of.}   %表达更加直接点。因此，增加正确的标注scribble是一个明智的选择。
%%%%%%%%%%%%%%%%%%%%%%%%%

\section{Related Work}
\label{sec::related_work}

\subsection{Salient Object Detection}
Traditional unsupervised works mainly achieve SOD based on low-level feature together with hand-picked priors such as global contrast priors~\cite{cheng2014global}, center priors~\cite{goferman2011context,jiang2013submodular, jiang2013salient}, and background priors~\cite{zhu2014saliency}. 
%\textbf{pseudo methods...}
Despite their simplicity, these methods cannot obtain important deep contextual semantic information, which results in poor performance.
Many methods~\cite{zhang2017supervision, zhang2018deep, nguyen2019deepusps, zhang2020learning, lin2022causal} exploit these handcrafted features as pseudo annotations to train a deep model. 
Nevertheless, they still rely on noise-based annotations, so achieving remarkable performance gains is challenging for them.
In recent years, deep learning has greatly driven the development of SOD. Majority approaches~\cite{liu2021visual,9deng2018r3net,13wang2019iterative,17wei2019f3net,xu2021locate,8chen2018reverse,12zhao2019pyramid,14liu2018picanet,15zhang2018bi,20pang2020multi,16chen2020global,22zhang2017amulet,23qin2019basnet,24feng2019attentive, zhao2019egnet} enhance the saliency detection model by improving network structure, such as iterative refining~\cite{9deng2018r3net,13wang2019iterative,17wei2019f3net,xu2021locate}, introduction of attention mechanism~\cite{8chen2018reverse,12zhao2019pyramid,14liu2018picanet,wang2019salient,li2021dense}, and researching fusion strategies~\cite{15zhang2018bi,20pang2020multi,16chen2020global,sun2021ampnet}. 
Some methods~\cite{22zhang2017amulet,23qin2019basnet,24feng2019attentive, zhao2019egnet, hui2023multi, tu2020edge} mainly utilize boundary-related information that contains abundant detail cues to help the model predict more accurate segmentation results.
In addition, \cite{wang2019inferring} introduces a fixation prediction task to reduce the semantic bias between the model and human visual mechanisms to accurately identify salient objects.
While these approaches have achieved great progress, they rely heavily on a large number of expensive pixel-wise labels.

%To address the limitations of low-level features, \textbf{several recent methods}~ have attempted to replace these low-level features with self-supervised learning (SSL) \textbf{features} for SOD. These SSL features are general image representations generated by learning on large unlabeled datasets, so they contain rich contextual information about the relationships between each region of the image. 
%Although the introduction of SSL features into unsupervised SOD is promising and has made some progress, the gap between these unsupervised methods and fully supervised SOD methods remains significant. In particular, it is difficult for them to distinguish which objects are salient, especially in complex scenes. \textit{The reason is that these unsupervised methods do not employ any strong manual annotation and therefore have to still rely on hand-picked priors with limited coverage such as \textbf{A, B, and C}.}

\subsection{Weakly Supervised Salient Object Detection}
\begin{figure}[t]
	\centering
	\includegraphics[width=0.48\textwidth]{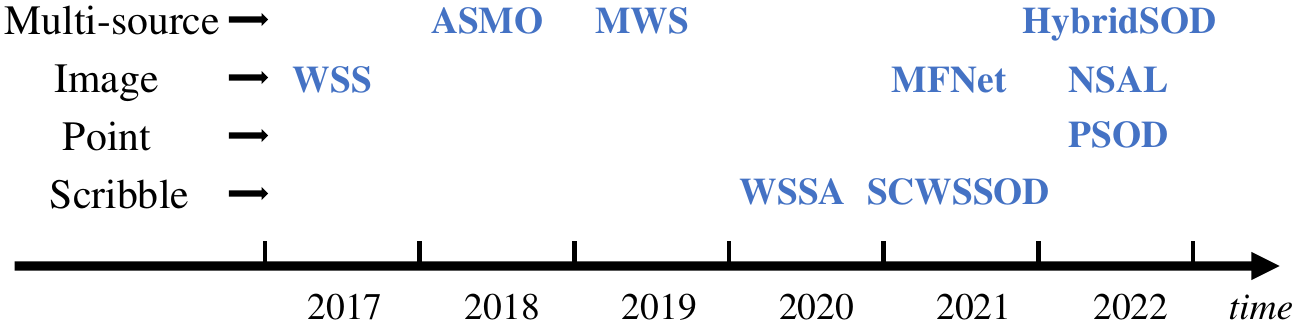} 
	\caption{A brief chronology of weakly supervised SOD methods. It consists of image-based (WSS~\cite{wang2017learning}, MFNet~\cite{piao2021mfnet}, and NSAL~\cite{ piao2022noise}), point-based (PSOD~\cite{gao2022weakly}), scribble-based (WAAS~\cite{zhang2020weakly} and SCWSSOD~\cite{yu2021structure}), and multi-source-based (AMSO~\cite{li2018weakly}, MWS~\cite{zeng2019multi}, and HybridSOD~\cite{cong2022weakly}) SOD methods.}
	\label{fig:time}
\end{figure}

With the rapid development of weakly supervised techniques, weakly supervised SOD methods, for example, image level-based, point-based, scribble-based, and multi-source-based are also explored to reduce the burden of pixel-wise annotation while maintaining model performance. Among them, the most widely researched is image-level-based SOD~\cite{wang2017learning, li2018weakly, piao2022noise, piao2021mfnet, zeng2019multi, cong2022weakly}.
Fig.~\ref{fig:time} shows a brief chronology.
\cite{wang2017learning} finds that using the classification network can extract foreground information related to salient objects, and then train the saliency detection network based on extracted pseudo labels.
Then,~\cite{piao2021mfnet, piao2022noise} note that the initial pseudo saliency map obtained by the classification network inevitably contains noise. Thus,~\cite{piao2021mfnet} filters out more precise maps from multiple noise pseudo labels based on multiple directive filters, and~\cite{piao2022noise} presents a noise-sensitive training strategy to balance the learning of object information and noise.
In addition, other methods~\cite{li2018weakly, zeng2019multi, cong2022weakly} explore multi-source weak labels that combine image-level labels with other weak annotations or pseudo tags generated by unsupervised manual methods to provide more comprehensive information to the model. 
However, these pseudo labels still contain rich noise that is difficult to correct in iterative training.
%Moreover, it is somewhat controversial to treat the highlight area calculated based on the classification network as the salient region generated by the human eyes.  
Point-supervised labels~\cite{gao2022weakly} can provide local ground truth and are generated with lower labeling time cost. Although only a few points are annotated, these annotations are obtained based on the human vision system, which can correctly guide the model without introducing any noise. But it is difficult for the model to discern the complete salient objects based a few points, especially for the boundary of the salient objects. Therefore,~\cite{gao2022weakly} uses an extra edge detector trained on edge datasets to help the model perceive the boundaries of the object and predict integral salient objects.

Scribble tags~\cite{zhang2020weakly} are a relatively cost-effective choice because they cover more regions of the salient objects and background, and do not consume much time compared with point annotations. Accordingly, this paper focuses on the scribble-based SOD method. \cite{zhang2020weakly} proposes the first scribble-based SOD method that applies an auxiliary edge detection task to help the model locate object edges. However, this work introduces an additional edge detector and does not fundamentally address the issue that the scribble tag itself lacks edge information. Furthermore, \cite{yu2021structure} explores the intrinsic properties of images and presents an LSC loss to improve local details. Although considerable progress has been achieved on scribble-based SOD, incomplete objects and noisy boundaries leave much room for improvement.

\subsection{Object Segmentation Properties of Self-supervised Vision Transformer}
With the emergence of self-supervised representation learning methods and the revelation of object characteristics, several unsupervised object detection methods based on self-supervised ViTs have been proposed.
Concretely,~\cite{simeoni2021localizing} proposes to select a seed patch from self-supervised representations likely to belong to a foreground object, and then expands more patches that have high similarity with the seed patch to discover the whole foreground object.
%Concretely, LOST~\cite{simeoni2021localizing} proposes to selects a seed patch from SSL features that is likely to belong to a foreground object, and then expands more patches that have high similarity with the seed patch to discovery the whole foreground object.
Then, TokenCut~\cite{wang2022self} presents a graph-based approach based on self-supervised ViT for unsupervised object discovery and extends this idea to unsupervised SOD. 
%Then, TokenCut~\cite{wang2022self} proposes a graph-based approach on the basis of SSL features for unsupervised object discovery and this idea is further extended to unsupervised SOD. 
Different from TokenCut,~\cite{shin2022unsupervised} proposes a spectral clustering method based on various self-supervised models for unsupervised SOD.
%Different from TokenCut, SELF-MASK~\cite{shin2022unsupervised} proposes a spectral clustering method based on various self-supervised models for unsupervised SOD.
Although the introduction of self-supervised representations into unsupervised SOD is promising and has made some progress, it is difficult for them to distinguish which objects are salient, especially in complex scenes, because they still depend on hand-crafted priors to determine salient objects.
Similar to image-level-based SOD, it is currently somewhat controversial to take the highlight area only based on self-supervised representations as the salient region generated by the human eyes.
Moreover, the gap between these unsupervised methods and fully supervised SOD methods remains significant.
Unlike the above methods that directly utilize self-supervised representations, this paper treats these features as general knowledge with rich object and scene information, and integrates them into the network to assist the model to perceive contextual affinities.
%\textbf{The results and contribution.}

%基于scribble的方法目前还是很少有工作。

\begin{figure*}[t]
	\centering
	\includegraphics[width=0.9\textwidth]{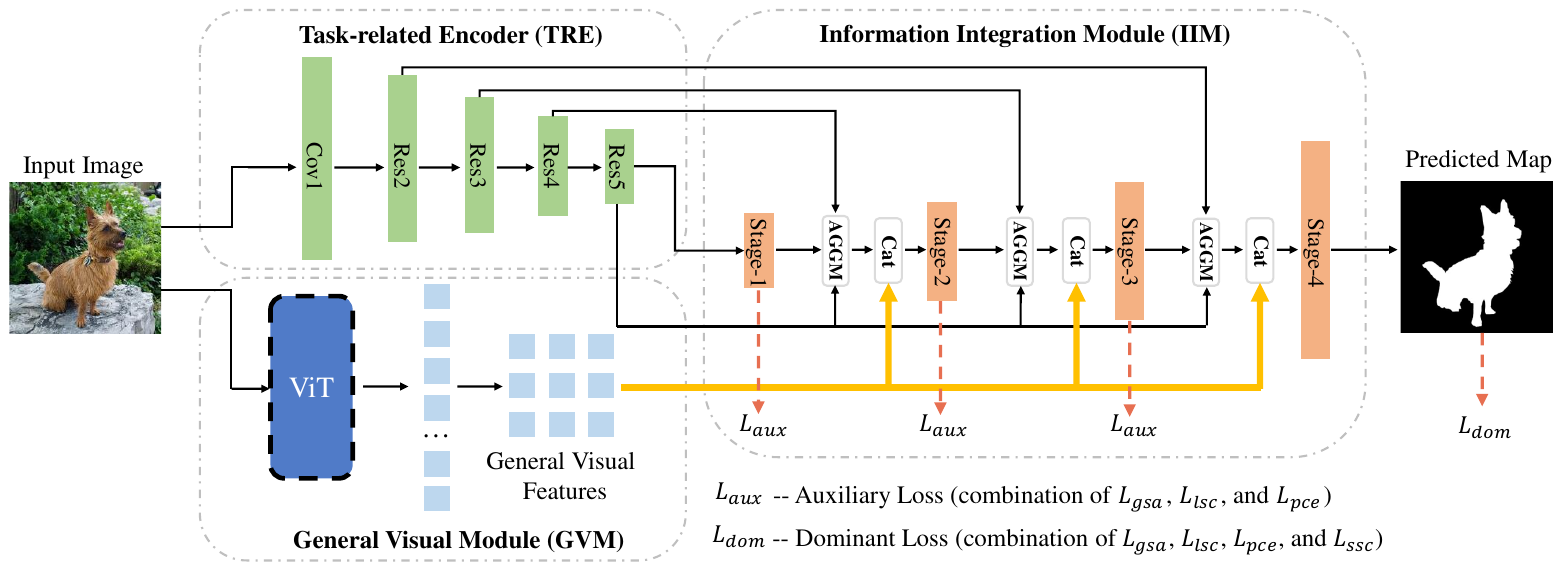} 
	\caption{
		Overview of our proposed framework. 
		It consists of a task-related encoder, a general visual module, and an information integration module. 
		GVM employs a ViT~\cite{dosovitskiy2020image} trained by self-supervised learning on a large-scale unlabeled datasets to provide general visual features.
		Our proposed GSA loss ($L_{gsa}$) is applied with LSC loss ($L_{lsc}$), saliency structure consistency loss ($L_{ssc}$), and partial cross entropy loss ($L_{pce}$) to guide the model. To capture the complete structure of salient objects, our $L_{gsa}$ guides the model to construct global affinities via maintaining the high semantic similarities within the salient objects and within backgrounds.
		%XXX loss XXX.
		%GIB trained on original images aims to identify integral salient objects, while BAB trained on synthetic images aims to help predict accurate boundaries. LSC loss $L_{lsc}$ is applied with saliency structure consistency (SSC) loss $L_{ssc}$ and partial cross entropy loss $L_{pce}$ to optimize the global integral branch. 
		%Local saliency coherence loss $L_{lsc}$ and partial cross entropy loss $L_{pce}$ are applied to optimize the boundary-aware branch. Self-consistent loss is employed to associate them.
	}
	\label{fig:framework}
\end{figure*}

\section{Methodology}
\label{sec::methodology}

%\subsection{Vision Transformers}
%Given an image of spatial size $H \times W$, vision transformers (ViT)~\cite{dosovitskiy2020image} use non-overlapping patches with resolution $K \times K$ as inputs and the total number of image patches is $HW/K^2$. Each patch is embedded in a numerical feature vector, represented as a token. 
%An extra learned token, called class token CLS, is designed to represent summary information for all image patches. 
%ViT consists of multiple layers of encoders, each with multi-head self-attention layers and feed-forward networks.
%Here, we utilize a vision transformer trained in a self-supervised manner based on DINO~\cite{caron2021emerging}, which is trained on ImageNet~\cite{deng2009imagenet} dataset without introducing any labels, and take latent variables from the final self-attention layer as \textbf{the global input features}.
%%ImageNet is a large visual database with millions of images, 
%Training on the ImageNet, a large visual database with millions of images, the model can learn general fundamental visual representations.
%\textbf{so that, need more words to support my opinion and contribution.}
%这些特征蕴含着丰富的知识，能够模拟人类的知识库。

\subsection{Framework}
%需要详细的阐述和论证
Our visual representation-guided framework contains a task-related encoder (TRE), a general visual module (GVM), and an information integration module (IIM), as shown in Fig.~\ref{fig:framework}.
Specifically, TRE is the common encoder that aims to filter irrelevant information and extract essential features to enforce the SOD task. 
%在训练中要反向传播和更新参数，包含不同分辨率和深度的特征
GVM primarily employs the ViT~\cite{dosovitskiy2020image} trained by self-supervised learning on a large-scale unlabeled datasets to provide general visual representations, which are like a wide range of human experience, thus helping the model understand the whole image and contextual relevance.
IIM gradually integrates various layers of task-related features and combines them with general visual features in each layers to realize their complementary advantages fully.
%the parameters of ViT are fixed during training
During training, the parameters of ViT are fixed, and it mainly provides general visual features to relieve the difficulty that task-related features cannot catch the contextual affinity of the objects or background in scribble-based SOD task, that is, the scribble label covers only part of the object area, making perceiving the structure of the object difficult for the model. 
%It helps the model understand the image from a global perspective and then, based on that, achieve more accurate saliency detection.
Meanwhile, TRE is designed to meet the requirements of the SOD task. Task-related features with various resolutions can compensate for the low resolution of the general visual features and guide the model to produce refined segmenting results.
During training, our proposed GSA loss is applied with LSC loss, the saliency structure consistency (SSC) loss~\cite{yu2021structure}, and the partial cross entropy loss as the dominant loss to guide the model. 
Simultaneously, we implement an auxiliary loss, which consists of the GSA loss, the LSC loss, and the partial cross entropy loss on each substage of IIM to supervise the intermediate predictions.

%Meanwhile, general visual features can relieve the difficulty that task-related features cannot catch the contextual affinity of the objects or background in scribble-based SOD task, i.e., scribble label covers only part of the object area, making it difficult for the model to perceive the structure of the object. 

%It helps the model to understand the image from a global perspective and then, based on that, achieve more accurate saliency detection.

\paragraph{Task-related Encoder}
TRE utilizes the encoder of SCWSSOD with backbone of ResNet-50\cite{he2016deep} pretrained on the ImageNet~\cite{deng2009imagenet} as the network structure, which can export multi-scale features with abundant information. It is composed of a head convolution and four residual layers. We select output features of four residual layers as final task-related features to feed into information integration module.

\paragraph{General Visual Module}
Given an image of spatial size $H \times W$, ViT uses non-overlapping patches with resolution $K \times K$ as inputs and the total number of image patches is $HW/K^2$. Each patch is embedded in a numerical feature vector, represented as a token. 
%An extra learned token, called class token CLS, is designed to represent summary information for all image patches. 
%ViT consists of multiple layers of encoders, each with multi-head self-attention layers and feed-forward networks.
We utilize the ViT trained in a self-supervised manner based on self-distillation loss (DINO)~\cite{caron2021emerging}, which is learned on large-scale dataset without introducing any labels, and take latent variables from the final self-attention layer as general vision representations. 
%Before transferring to IIM, we reshape the general vision representations to $H/K \times W/K$.
Before concatenating with the corresponding features of IIM, we normalize general visual representations and resize them to the same size as these features by reshape operations and convolution layers. 

\begin{figure}[t]
	\centering
	\includegraphics[width=0.48\textwidth]{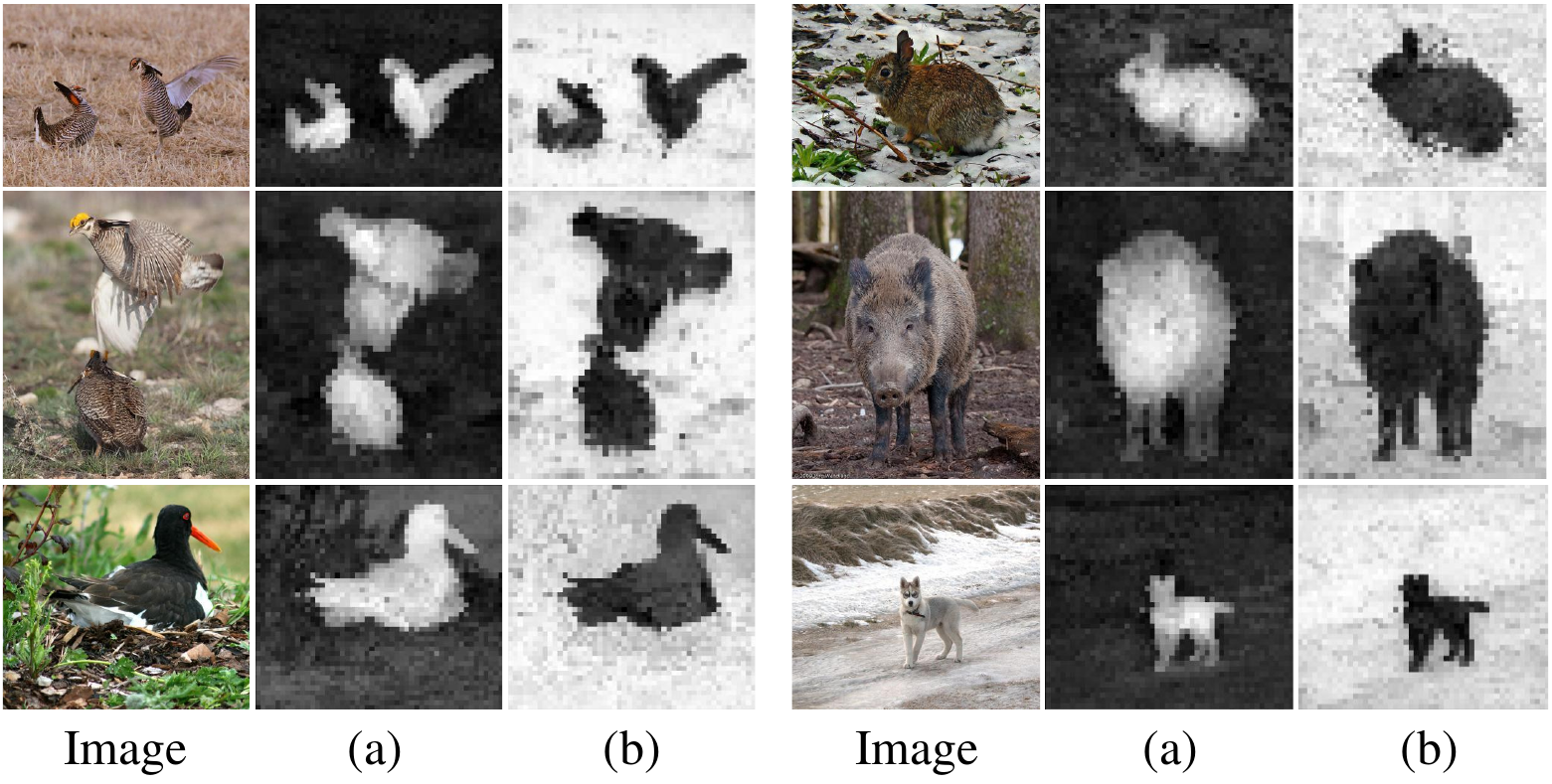} 
	\caption{Visual results of the scribble similarity maps. (a) denotes the foreground scribble similarity map and (b) denotes background scribble similarity map.}
	\label{fig:scribble_similarity_map}
\end{figure}

To demonstrate that general visual representations possess rich contextual semantic information, we sample six challenging images and present their foreground scribble similarity maps and background scribble similarity maps, as shown in Fig.~\ref{fig:scribble_similarity_map}. The process of generating these similarity maps is relatively simple. By calculating the cosine distance between two points in general visual representations, the similarity value between the two points can be obtained. Then, we can get foreground scribble similarity map by calculating the average similarity between each point and all foreground scribble points. 
The highlighted regions in the foreground scribble similarity maps indicate a strong correlation with the foreground scribble. 
The process of generating the background scribble similarity maps is similar. 
These sampled images contain various challenging scenes, such as complex backgrounds, high foreground-background similarity, multiple objects, and multi-colored foreground.
However, we can observe that the objects in both the foreground and background scribble similarity maps can be completely and accurately presented.
This phenomenon indicates that, given an image and the corresponding scribble tags, it is feasible to accurately distinguish which parts of the image belong to the same object as the scribble labels by utilizing general visual representations. Thus, general visual representations with rich contextual semantic information can help the model capture more complete salient objects.

%ImageNet is a large visual database with millions of images, 
%\textbf{Training on the ImageNet, a large visual database with millions of images, the model can learn general fundamental visual representations.}
%\textbf{so that, need more words to support my opinion and contribution.}

\paragraph{Information Integration Module}
IIM not only needs to combine task-related features and general visual features to obtain comprehensive information, but also recover the spatial size of features, that is, generate high-resolution results by merging high and low-level features. The basic structure of this decoder is the same as those of SCWSSOD, where aggregation module (AGGM) can integrate multi-level features and learn the weight of each feature. We additionally incorporate general visual features into the decoder through a simple concatenate operation.
The excellent experimental results (Section~\ref{sec::experiment}) show that such simple concatenate operation can effectively integrate general visual representation into the whole framework.

\subsection{Global Semantic Affinity Loss}
%引入通用特征能够有效提升模型的辨别能力，但是对于弱监督学习来说，制定合适的损失函数让模型逼近任务要求也是至关重要的。
%scribble标签位于显著性目标或者背景的内部，能够提供显著性目标的定位信息。但是缺乏其他区域，尤其是边缘部分的标注，所以该任务的难点之一就是如何构建全局的上下文联系，引导模型感知目标的整体结构。
%传统的分割方法normalized cut考虑了 popular graph clustering algorithm，
The introduction of general visual representation can effectively improve the discrimination of the model, but for weakly supervised learning, it is also crucial to develop an appropriate loss function to approach the task requirements.
The scribble label is located inside the salient object and background, so it can provide the concrete location information of the salient object. 
Nevertheless, how to construct global contextual affinity constraints to guide the model to perceive the structure of the object is a pending challenge because most region annotations are absent.

%We have discovered that such a projection can be used with Normalized Cut [43] (Ncut) to significantly improve foreground / background segmentation.
A classic graph clustering segmentation algorithm Normalized Cut (Ncut)~\cite{shi2000normalized} considers the global contextual affinity of the image and divides a graph into two non-overlapping sets. 
This method constructs an Ncut energy and minimizes it to maintain the low similarity between sets, which is defined as follows:
\begin{equation*}
Ncut\left( A,B\right) =\dfrac{C\left( A,B\right) }{C\left( A,V\right) }+\dfrac{C\left( A,B\right) }{C\left( B,V\right) },
\end{equation*}
where $C$ indicates the degree of similarity between two sets. $C\left( A,B\right) = \sum _{v_i\in A,v_{j}\in B}\varepsilon _{i,j}$, where $\varepsilon _{i,j}$ denotes the similarity value of node $v_i$ and node $v_{j}$. $C\left( A,V\right)$ measures the degree of similarity between nodes in $A$ and all nodes in the graph.
%最近，XXX等方法将Ncut作为正则化损失函数引入分割的任务中，并且验证了其有效性。该方法基于低层的RGBXY空间构建节点间的相似性，对于相当简单的图片可能有一定效果，但是deal with the 稍微 complex situation of low foreground/background contrast.
%Recently,~\cite{tang2018normalized} introduces Ncut as a regularized loss for the task of segmentation and the experimental results verified its validity.
This method builds the similarity between nodes over the low-level color space, which may be effective to deal with the simple images, but is difficult for the complex scene of low foreground/background contrast.
%但是基于通用特征能够构建上下文联系。

\begin{figure}[t]
	\centering
	\includegraphics[width=0.48\textwidth]{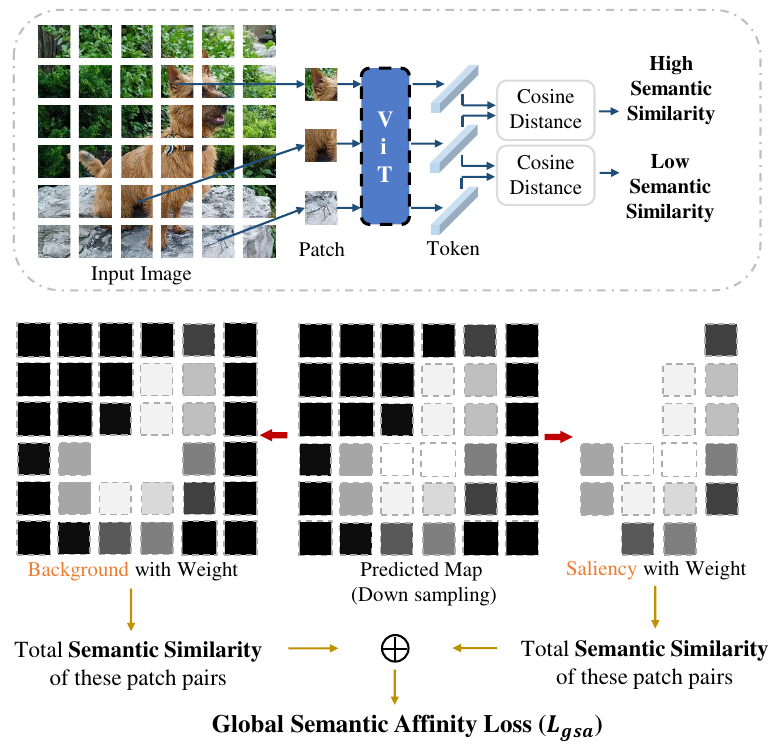} 
	\caption{Illustration of GSA loss, whose main idea is to maintain high similarity among nodes within the same class by introducing semantic cosine distance.
	}
	\label{fig:gsa_loss}
\end{figure}

Instead of color feature, we use general visual representation based on DINO, which exhibits a significant ability to establish correlations between objects even in complex cases, to construct semantic similarities between nodes.
In addition, unlike directly classifying the nodes in the graph by the minimum cut algorithm, we design the GSA loss based on the idea of Ncut to guide the model to focus on global contextual consistency.
%Concretely, the idea of GSA is to max the similarity in the same set to achieve the goal of maintaining the high similarity of nodes in the same set and the low similarity between sets, which is similar to that of Ncut.
Concretely, the idea of GSA is to max the similarity in the same set to achieve the goal of maintaining the high similarity of nodes in the same set, which is similar to that of Ncut.
It is defined as follows:
\begin{equation*}
GSA(A,B) =\dfrac{C\left( A,A\right) }{C\left( A,V\right) }+\dfrac{C\left( B,B\right) }{C\left( B,V\right) }.
\end{equation*}
%因为我们设计的模型预测的结果并不是二分类的

Specifically, similar to the GVM, the images are cut into non-overlapping patches, and then the visual semantic features of each patch are obtained through ViT. 
We construct the semantic similarity between the patches by cosine distance, as shown in Fig.~\ref{fig:gsa_loss}.
In addition, Ncut needs to compute the similarity between nodes in different sets, whereas the idea of the GSA loss computes the similarity between patches in the same set.
The reason is that calculating inter-class similarity is required to divide all patches into two classes, whereas the patch predictions of our model can be any value between 0, 1. 
Hence, we tackle this problem by calculating intra-class similarity.
We can treat these patches as saliency and background points.
For example, a patch with a predicted value of 0.3 can be considered as a point of the salient object with a weight of 0.3 and a background point with a weight of 0.7.
Finally, we set a global energy to maintain the high similarity of patches in the same classification.
%It can also be understood that two pixels i and j of down-sampling predicted map to draw close saliency scores for corresponding patches with similar features.
The GSA loss of saliency part $L_{gsa}^f$ is defined as follows:
\begin{equation}
L_{gsa}^f =1 - \dfrac{\sum\limits _{i\in R}\sum\limits _{j\in R}s_i\cdot s_j\cdot CS\left( t_{i},t_{j}\right) }{\sum\limits _{i\in R}\sum\limits _{j\in R}s_{i}\cdot CS\left( t_i,t_{j}\right) },
\end{equation}
where $R$ is a set of image patches, $s_i$ represents the predicted saliency value of patch $i$, and $t_i$ indicates the semantic features of the token $i$. $CS(\cdot,\cdot)$ denotes the cosine similarity, which is expressed as: 
\begin{equation}
CS(p,q)= \dfrac{p\cdot q}{\left\| p\right\| _{2}\cdot \left\| q\right\| _{2}}.
\end{equation}
The GSA loss of background part $L_{gsa}^b$ is defined as follows:
\begin{equation}
L_{gsa}^b =1 - \dfrac{\sum\limits _{i\in R}\sum\limits _{j\in R}(1-s_i)\cdot (1-s_j)\cdot CS\left( t_{i},t_{j}\right) }{\sum\limits _{i\in R}\sum\limits _{j\in R}(1-s_i)\cdot CS\left( t_i,t_{j}\right) }.
\end{equation}
The final GSA loss $L_{gsa}$ is defined as follows:
\begin{equation}
L_{gsa} = L_{gsa}^f + L_{gsa}^b.
\end{equation}

\subsection{Loss Function}
Based on scribble supervision, local correlation, and scale consistency, we propose global affinity to guide the model to capture the intact salient object. The training loss $L$ is defined as follows:
\begin{equation}
L = L_{dom} + \sum ^{3}_{k=1}\lambda _{k}L_{aux}^{k},
\end{equation}
where $\lambda _{k}$ is set to balance the weight of each stage and we take the same weights as in GCPANet~\cite{16chen2020global}. 
The dominant loss $L_{dom}$ and auxiliary loss $L_{aux}$ are formulated as follows:
\begin{equation}
L_{dom} = L_{pce} + L_{ssc} + \beta L_{lsc} + \mu L_{gsa},
\end{equation}
\begin{equation}
L_{aux}^{k} = L_{pce} + \beta L_{lsc} + \mu L_{gsa} \qquad  k\in\left\{ 1,2,3\right\},
\end{equation}
where $\beta$ is used to adjust the weight of the LSC loss and it shares the same weight 0.3 as SCWSSOD. $\mu$ is applied to adjust the weight of the GSA loss.

%\begin{equation}
%L = L_{pce} + L_{ssc} + L_{lsc} + \alpha L_{good}.
%\end{equation}

$L_{pce}$ is the partial cross entropy loss, which can be written as follows:
\begin{equation}
L_{pce}=\sum _{i\in J}y_{i}\log \widehat{y}_{i}-\left( 1-y_{i}\right) \log \left( 1-\widehat{y}_{i}\right).
\end{equation}

The SSC loss $L_{ssc}$~\cite{yu2021structure} is introduced to strengthen model's generalization ability by self-supervised learning with different image scales, which is expressed as follows:
\begin{equation}
L_{ssc} = \alpha \dfrac{1-SSIM\left( S^{\Downarrow },S^{\downarrow } \right) }{2}+\left( 1-\alpha \right) \left| S^{\Downarrow }-S^{\downarrow } \right|,
\end{equation}
where $SSIM$ is the single scale structural similarity (SSIM)~\cite{wang2004image, godard2017unsupervised} and $\alpha$ is set to 0.85. $S^{\downarrow }$ represents down-scaled saliency map, and $S^{\Downarrow }$ represents the saliency map of the same image with down-scaled size.

LSC loss~\cite{yu2021structure} penalizes pixels with the similar color in local regions, which
is defined as follows:
\begin{equation}
L_{lsc}=\sum _{i}\sum _{j\in K_{i}}F\left( i,j\right) D\left( i,j\right),
\end{equation}
where $D\left( i,j\right) =\left| S_{i}-S_{j}\right|$ denotes the saliency difference between the pixel $i$ and the pixel $j$ and $K_i$ is the local area. $F(i,j)$ is the similarity energy that assigns similar saliency scores for proximal pixels with similar color~\cite{obukhov2019gated}, which is formulated as follows:
\begin{equation}
F\left(i,j\right) =\dfrac{1}{\omega }\exp \left(-\dfrac{\left\| I\left( i\right) -I\left( j\right) \right\| ^{2}}{2\sigma _{I}^{2}}-\dfrac{\left\| P\left( i\right) -P\left( j\right) \right\| ^{2}}{2\sigma _P^{2}}\right),
\end{equation}
$I(\cdot)$ and $P(\cdot)$ represent the RGB value and position of a pixel, respectively; $1/{\omega }$ denotes the normalized weight; $\sigma _{I}$ and $\sigma _P$ denotes the Gaussian kernel scale, and $\left\| \cdot \right\|$ is an $L2$ operation. $\omega$, $\sigma _{I}$, and $\sigma _P$ are set to 1, 6 and 0.1, respectively. They shares the same weight as SCWSSOD.

\section{Experiments}
\label{sec::experiment}

\subsection{Datasets and Implementation Details}
\subsubsection{Datasets and Evaluation Metrics}
Our method is trained on scribble labeled dataset S-DUTS~\cite{zhang2020weakly}, which contains 10553 images. To validate the quality of our method, we conduct experiments on five SOD benchmarks, DUTS-TE~\cite{wang2017learning}, PASCAL-S~\cite{li2014secrets}, ECSSD~\cite{2yan2013hierarchical}, HKU-IS~\cite{li2015visual}, and DUT-OMRON~\cite{yang2013saliency}.
We adopt three widely-used evaluation metrics: F-measure~\cite{achanta2009frequency}, MAE, and E-measure ($E_m$)~\cite{fan2018enhanced}. We employ the max f-measure over all thresholds, denoted as $F_{max}$. $F_{avg}$ denotes the mean F-measure.
\subsubsection{Implementation Details}
We conduct our method on an NVIDIA GeForce GTX 3090 with PyTorch.
Input images are resized to 320$\times$320 and the size of the initial general visual features is $40\times40\times384$.
We use horizontal flips for data augmentation.
In the training phase, we employ the stochastic gradient descent (SGD) optimizer with a triangular warm-up and decay strategy with the maximum learning rate of 5e-3 and the minimum learning of 1e-5, a momentum of 0.9, and a weight decay of 5e-4. The total epoch is 40 and the batch size is set to 16.
These hyperparameters are same as SCWSSOD.
We employ the ViT-S/8 model~\cite{dosovitskiy2020image} trained with DINO to extract general visual features of patches.

The general visual features of each image are repeatedly utilized during training, so we firstly generate and save these features before training and then directly employ them to speed up the training. These features can be directly used for GSA loss.
No post-processing is used.
Our model contains around 69M trainable parameters and runs around 21 FPS for inference on an NVIDIA GeForce GTX 3090.
It requires a memory footprint of around 16 GB, and the training process takes about 11 hours.
%The training time is about 11 hours and the memory footprint required for this is around 14GB.

%\textbf{S/16, VIT
%The “-S” and “-B” are ViT small[6, 18] and ViT base[6, 18] architecture respectively. The “-16” and “-8” represents patch sizes 16 and 8 respectively.}

\begin{table*}[t!]
	%\small
	%\footnotesize 
	\scriptsize
	\caption{Quantitative comparisons with seven state-of-the-art weakly supervised/unsupervised methods and nine state-of-the-art fully supervised methods on five datasets in terms of max F-measure ($F_{max}$), mean F-measure ($F_{avg}$), MAE ($MAE$), and E-measure ($E_m$). The best results are marked in bold. ``Sup.'' means supervision information. ``Fully'' means fully supervised. ``Unsup.'' means unsupervised. ``Image'' means image-level supervised. ``Point'' means point-level supervised. ``scribble'' means scribble-level supervised. ``Multi'' means multi-source supervised.}
	%\tiny
	\begin{center}
		\setlength{\tabcolsep}{0.4mm}{
			\begin{tabular}{c|c|cccc|cccc|cccc|cccc|cccc}
				\toprule
				\multirow{2}*{Methods}& \multirow{2}*{Sup.}& \multicolumn{4}{c|}{DUTS-TE} & \multicolumn{4}{c|}{PASCAL-S} & \multicolumn{4}{c|}{ECSSD} & \multicolumn{4}{c|}{HKU-IS} & \multicolumn{4}{c}{DUT-OMRON}\\ 
				
				%\cline{2-21}
				%\cmidrule(l){2-5} \cmidrule(l){6-9} \cmidrule(l){10-13} 
				
				&
				%&\multicolumn{3}{}{method}
				&\multicolumn{1}{c}{$F_{max}$}&{$F_{avg}$}&{$MAE$}&{$E_m$}	
				&\multicolumn{1}{c}{$F_{max}$}&{$F_{avg}$}&{$MAE$}&{$E_m$}
				&\multicolumn{1}{c}{$F_{max}$}&{$F_{avg}$}&{$MAE$}&{$E_m$}	 	 
				&\multicolumn{1}{c}{$F_{max}$}&{$F_{avg}$}&{$MAE$}&{$E_m$}	 
				&\multicolumn{1}{c}{$F_{max}$}&{$F_{avg}$}&{$MAE$}&{$E_m$}	 
				
%				&\multicolumn{1}{c}{$F_{max} \uparrow$}&{$F_{avg} \uparrow$}&{$M \downarrow$}&{$E_m \uparrow$}	
%				&\multicolumn{1}{c}{$F_{max} \uparrow$}&{$F_{avg} \uparrow$}&{$M \downarrow$}&{$E_m \uparrow$}
%				&\multicolumn{1}{c}{$F_{max} \uparrow$}&{$F_{avg} \uparrow$}&{$M \downarrow$}&{$E_m \uparrow$}	 	 
%				&\multicolumn{1}{c}{$F_{max} \uparrow$}&{$F_{avg} \uparrow$}&{$M \downarrow$}&{$E_m \uparrow$}	 
%				&\multicolumn{1}{c}{$F_{max} \uparrow$}&{$F_{avg} \uparrow$}&{$M \downarrow$}&{$E_m \uparrow$}	 
				\\ 
				\midrule
				
				%PAKRN (2021)~\cite{xu2021locate}      & F & .852 & \textbf{.838} & .066 & \textbf{.857} & .928 & \textbf{.931} & \textbf{.032} & .924 & \textbf{.923} & \textbf{.920} & \textbf{.032} & \textbf{.955} & \textbf{.853} & \textbf{.793} & \textbf{.050} & \textbf{.885} & \textbf{.900} & \textbf{.865} & \textbf{.033}  & \textbf{.916}\\
				%SGLKRN (2021)~\cite{xu2021locate}     & F  & .830 & .067 & .859  & .922 & .036 & .927  & .916 & .028 & .954  & .783 & .049 & .883  & .851 & .034 &.913\\
				DGRL (2018)~\cite{wang2018detect}       & Fully    &.828& .794 & .050 &.879 &.848& .801 & .073 & .834  &.925& .903 & .043 & .917  &.913& .882 & .043 & .941  &.779& .709 & .063 & .843  \\ 
				MLMSNet (2019)~\cite{wu2019mutual}      & Fully    &.852& .745 & .049 &.860 &.850& .758 & .073 & .836  &.928& .869 & .045 & .914  &.921& .871 & .039 & .953  &.774& .692 & .064 & .865  \\
				BASNet (2019)~\cite{23qin2019basnet}    & Fully    &.860& .791 & .048 &.884 &.854& .771 & .075 & .846  &.942& .880 & .037 & .921  &.928& .895 & .032 & .946  &.805& .756 & .056 & .869  \\ 
				PoolNet (2019)~\cite{liu2019simple}     & Fully    &.880& .809 & .040 &.889 &.863& .815 & .074 & .848  &.944& .915 & .039 & .924  &.932& .899 & .033 & .948  &.808& .747 & .056 & .863  \\
				MINet (2020)~\cite{20pang2020multi}     & Fully    &.884& .828 & .037 &.898 &.867& .829 & .063 & .851  &.947& .924 & .033 & .927  &.935& .909 & \textbf{.029} & \textbf{.953}  &.810& .755 & .055 & .865  \\ 
				GateNet (2020)~\cite{zhao2020suppress}  & Fully    &.888& .807 & .040 &.889 &.869& .819 & .067 & .851  &.945& .916 & .040 & .924  &.933& .899 & .033 & .949  &.818& .746 & .055 & .862  \\
				GCPANet (2020)~\cite{16chen2020global}  & Fully    &.888& .817 & .038 &.891 &.869& .827 & \textbf{.061} & .847  &.948& .919 & .035 & .920  &.938& .898 & .031 & .949  &.812& .748 & .056 & .860  \\ 
				VST (2021)~\cite{liu2021visual}         & Fully    &.890& .818 & \textbf{.037} &.892 &.875& .829 & \textbf{.061} & .837  &\textbf{.951}& .920 & .033 & .918  &\textbf{.942}& .900 & \textbf{.029} & \textbf{.953}  &.825& .756 & .058 & .861  \\
				ICON-R (2022)~\cite{zhuge2022salient}   & Fully    &\textbf{.892}& .838 & \textbf{.037} &.902 &.876& .833 & .063 & .855  &.950& \textbf{.928} & \textbf{.032} & \textbf{.929}  &.939& .910 & \textbf{.029} & .952  &.825& .772 & .057 & .870  \\
				Mguid-Net (2023)~\cite{hui2023multi}    & Fully    &\textbf{.892}& .815 & \textbf{.037} &.897 &.879& .802 & \textbf{.061} & .852  &.946& .910 & .036 & .918  &.938& .897 & .031 & .951  &.805& .756 & .056 & .869  \\
				TokenCut (2022)~\cite{wang2022tokencut}	& Unsup.   &.796& .600 & .128 &.757 &.746& .668 & .181 & .745  &.833& .777 & .156 & .827  &.834& .743 & .133 & .835  &.793& .666 & .127 & .790  \\ 
				MWS (2019)~\cite{zeng2019multi}         & Multi    &.767& .685 & .091 &.814 &.784& .712 & .132 & .784  &.878& .840 & .096 & .884  &.856& .814 & .084 & .895  &.718& .609 & .109 & .763  \\ 
				MFNet (2021)~\cite{piao2021mfnet}       & Image    &.763& .693 & .079 &.830 &.797& .746 & .111 & .812  &.873& .844 & .084 & .887  &.875& .839 & .058 & .917  &.685& .621 & .098 & .783  \\
				NSAL (2022)~\cite{piao2022noise}        & Image    &.781& .730 & .073 &.849 &.795& .756 & .110 & .816  &.878& .856 & .078 & .883  &.882& .864 & .051 & .923  &.715& .648 & .088 & .801  \\ 
				PSOD (2022)~\cite{gao2022weakly}	    & Point    &.858& .812 & .045 &.887 &.866& .831 & .064 & .851  &.936& .921 & .036 & .924  &.923& .907 & .032 & .952  &.809& .748 & .064 & .854  \\ 
				WSSA (2020)~\cite{zhang2020weakly}      & Scribble &.789& .742 & .062 &.857 &.809& .774 & .092 & .831  &.888& .870 & .059 & .901  &.881& .860 & .047 & .927  &.753& .703 & .068 & .840  \\
				SCWSSOD (2021)~\cite{yu2021structure}   & Scribble &.844& .823 & .049 &.890 &.841& .818 & .077 & .846  &.915& .900 & .049 & .908  &.909& .896 & .038 & .938  &.783& .758 & .060 & .862  \\ 
				
				\midrule
				Ours                                    & Scribble  &.877& \textbf{.861} & \textbf{.037} &\textbf{.908} &\textbf{.891}& \textbf{.859} & .071 & \textbf{.877}  &.941& \textbf{.928} & .037 & .925  &.928& \textbf{.919} & .032 & .949  &\textbf{.892}& \textbf{.873} & \textbf{.033} & \textbf{.927} \\
				\bottomrule

		\end{tabular}}
	\end{center}
	
	\label{table:state_of_the_art}
\end{table*}

\begin{figure*}[t!]
	\centering
	\includegraphics[width=1
	\textwidth]{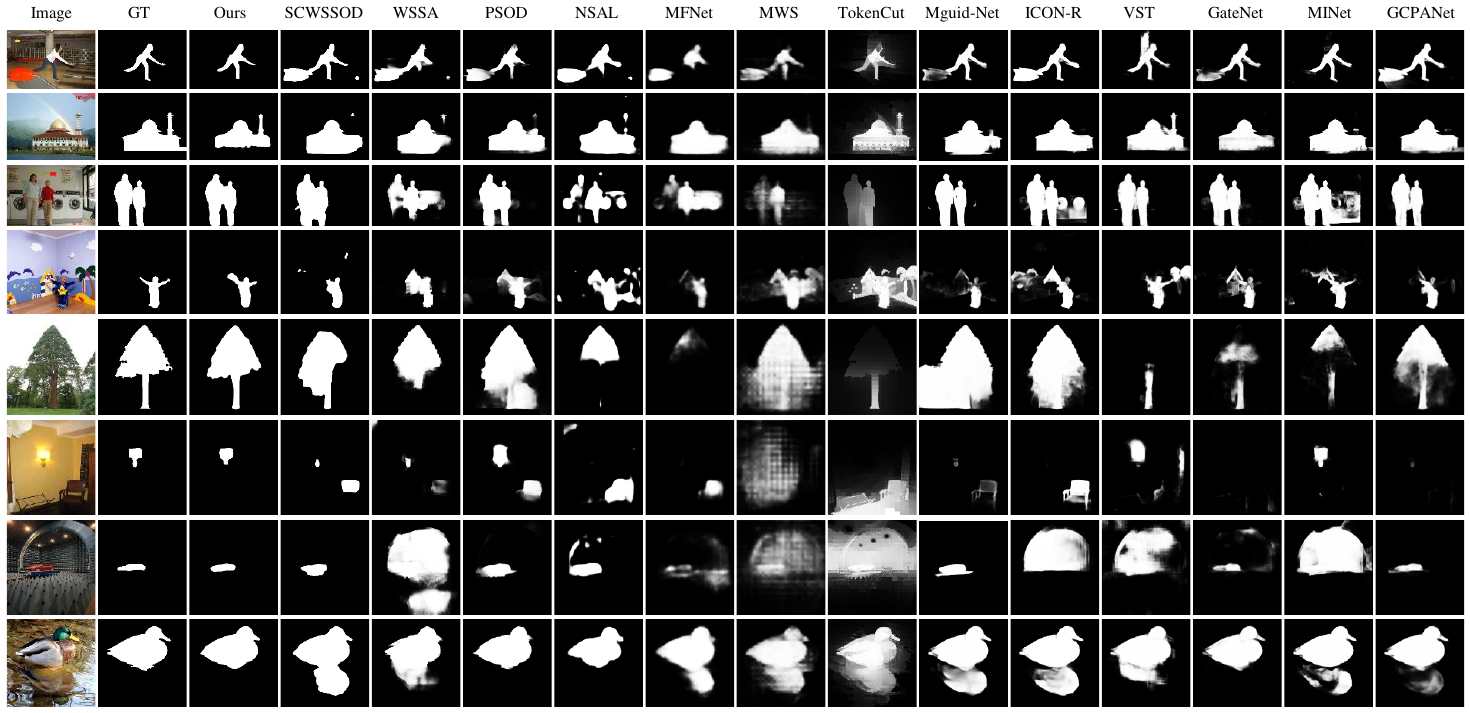}
	\caption{Visual comparisons with different methods. Each column represents an approach and each row shows the saliency map of an image. Apparently, our predicted saliency maps are more accurate and complete.}
	\label{fig:visual_map}
\end{figure*}

\subsection{Comparison with the State-of-the-arts}
\subsubsection{Quantitative Comparison}
We compare our methods with seven state-of-the-art weakly supervised/unsupervised methods (TokenCut (2022)~\cite{wang2022tokencut}, MWS (2019)~\cite{zeng2019multi}, MFNet (2021)~\cite{piao2021mfnet}, NSAL (2022)~\cite{piao2022noise}, PSOD (2022)~\cite{gao2022weakly}, WSSA (2020)~\cite{zhang2020weakly}, and SCWSSOD (2021)~\cite{yu2021structure}) and ten state-of-the-art fully supervised methods (DGRL (2018)~\cite{wang2018detect}, MLMSNet (2019)~\cite{wu2019mutual}, BASNet (2019)~\cite{23qin2019basnet}, PoolNet (2019)~\cite{liu2019simple}, MINet (2020)~\cite{20pang2020multi}, GateNet (2020)~\cite{zhao2020suppress}, GCPANet (2020)~\cite{16chen2020global}, VST (2021)~\cite{liu2021visual}, ICON-R (2022)~\cite{zhuge2022salient}, and Mguid-Net (2023)~\cite{hui2023multi}), as shown in Table~\ref{table:state_of_the_art}.
For fair comparisons, the saliency maps of these 17 methods are directly provided by the author or their released codes and we assess them with the same evaluation code.
The best results are marked in bold.
Compared with weakly supervised or unsupervised methods, our methods achieve a new state-of-the-state performance under all the metrics. 
Concretely, our method outperforms previous best scribble-based method (SCWSSOD) on all datasets by a large margin with an average gain of 5.5\% for $F_{max}$, an average gain of 5.8\% for $F_{avg}$, an average gain of 24\% for $MAE$, and an average gain of 3.1\% for $E_m$.
%Our methods achieves an average gain of 5.2\% for $F_\beta$, an average gain of 12.5\% for $MAE$, and an average gain of 2.6\% for $E_m$, compared with the best weakly supervised method (PSOD). PSOD employs an extra edge detector trained on edge datasets to enhance the model, 
What's more, our methods achieve an average gain of 1.0\% for $F_{max}$, an average gain of 3.7\% for $F_{avg}$, an average gain of 4.5\% for $MAE$, and an average gain of 1.7\% for $E_m$ compared with the best fully supervised method (ICON-R). 
Our method that only utilizes scribble annotations without introducing any extra label is comparable or even superior to the state-of-the-art fully supervised method, which demonstrates that our method is absolutely effective and robust.
In addition, our method is far superior to other methods on DUT-OMRON. 
For fully supervised method, even VST and ICON-R, the results on $F_{avg}$ are below 0.8, which shows that these models usually incorrectly distinguish salient objects or fails to identify the complete structure of the salient object on DUT-OMRON.
The great improvement of our method is that it can construct global contextual affinities to identify the structure of salient objects better.
However, on simple datasets, such as ECSSD and HKU-IS, our method does not have an absolute advantage. 
The reason is that, for these simple images where salient objects can be easily localized, and fully supervised approaches with pixel-wise information can handle the details better.

\subsubsection{Qualitative Comparison}
To further evaluate the advantages of our proposed method, we sample eight images from DUTS-TE~\cite{wang2017learning} and DUT-OMRON~\cite{yang2013saliency} and shows the saliency maps predicted by seven weakly supervised or unsupervised methods and six state-of-the-art fully supervised method in Fig.~\ref{fig:visual_map}.
%我们可以发现，在嘈杂的背景下，
It can be seen that our weakly supervised method can handle various challenging scenarios, including foreground disturbance, cluttered background, similar fore-background, and
multi-object interference.
Compared with previous weakly supervised methods or unsupervised methods, our predicted saliency maps are more accurate and complete.
Compared with fully supervised methods, although our method performs slightly worse in boundary details, our method can capture correct salient objects and predict more complete structure.
These results illustrate the immense potential of building SOD models based on general visual features and the robustness of the GSA loss, which can effectively help and guide the model to capture the global contextual affinities to recognize the complete salient object.
Additionally, we can see that image-level SOD methods (i.e., NSAL and MFNet) and the unsupervised method TokenCut based on hand-crafted priors have difficulty capturing real salient objects without any human annotation on image 5, 6, and 7.

\begin{table}
	%\small
	\centering
	\caption{Ablation study for general visual features and GSA loss on DUTS-TE and PASCAL-S. ``GVM'' denotes general visual module and ``GSA'' denotes the GSA loss.}
	\setlength{\tabcolsep}{1.5mm}{
		\begin{tabular}{c|c|c|ccc|ccc}
			%\multirow{2}{*}{Baseline}& \multirow{2}{*}{O}& \multirow{2}{*}{D} & \multirow{2}{*}{E} & \multicolumn{3}{|c|}  \\
			% &  &&  \\
			\toprule
			\multirow{2}*{Baseline}& \multirow{2}*{GVM}& \multirow{2}*{GSA}  & \multicolumn{3}{c|} {DUTS-TE}& \multicolumn{3}{c} {PASCAL-S}   \\
			&&&$F_{avg} $ &$MAE $ &$E_m $ &$F_{avg} $ &$MAE $ &$E_m $          \\
			%Baseline&SSL & Loss  &$F_{avg} \uparrow$ &$M \downarrow$ &$E_m \uparrow$ \\
			\midrule
			$\surd$&&&                     	   .823    & .049  &.890     &.818    & .077  &.846\\
			$\surd$&$\surd$&&                  .838    & .039  &.906     &.835    & .075  &.873\\
			$\surd$&&$\surd$&                  .831    & .046  &.893     &.826    & .073  &.852\\
			$\surd$&$\surd$&$\surd$&           .861    & .037  &.908     &.859    & .071  &.877\\
			\bottomrule
	\end{tabular}}
	\label{table:ablation_loss}
\end{table}

\begin{figure}[h]
	\centering
	\includegraphics[width=0.45\textwidth]{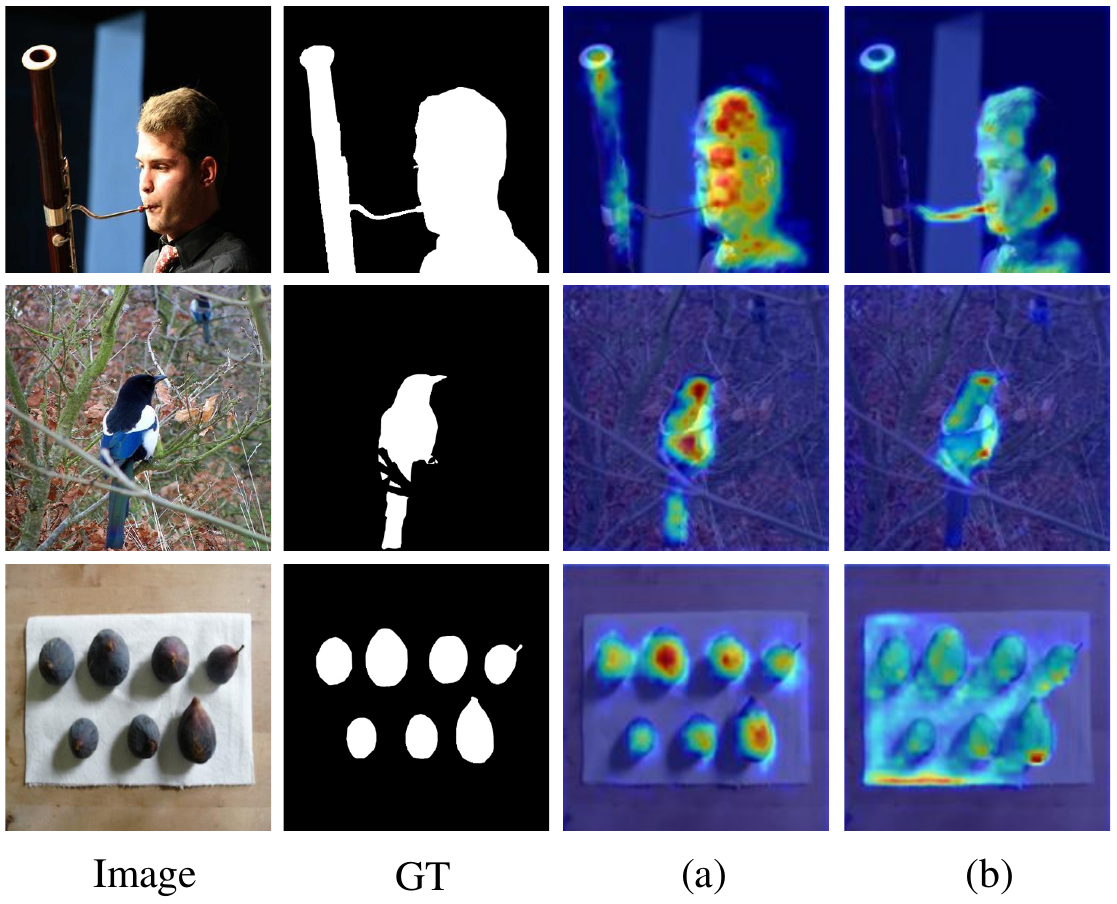} 
	\caption{Visual results of the feature map from the IIM. (a) denotes the feature maps after fusing general visual features in our network and (b) denotes the feature maps without fusing general visual features in the same layers of the baseline network.}
	\label{fig:feature_map}
\end{figure}

\subsection{Ablation Study}

\subsubsection{Effectiveness of General Visual Features}
%为了验证A模块的有效性，我们直接将我们的方法和single分支的方法进行比较。
%XXX表示XXX.

To evaluate the effectiveness of introducing general visual features, we conduct a series of comparisons with and without GVM on DUTS-TE and PASCAL-S datasets, as shown in Table~\ref{table:ablation_loss}.
We directly use SCWSSOD as the baseline model. 
As we can see, whether or not the GSA loss is applied, when general visual features is introduced, the results on all metrics are substantially improved.
This phenomenon proves that the idea of implementing SOD supported by general visual features is reliable and effective for excellent performance.

Fig.~\ref{fig:feature_map} provides some visual examples of the feature map from the stage-2 of IIM to illustrate the advantages of introducing general visual features. Fig.~\ref{fig:feature_map}(a) denotes the feature maps after fusing general visual features in our network and Fig.~\ref{fig:feature_map}(b) denotes the feature maps without fusing general visual features in the same layers of the baseline network.
We can see that our network perceives better long-distance connections in an image and precisely identifies which regions belong to the same object  such as recognizing complete instruments, detecting the non-adjacent bird tail, and distinguishing between fruits and their shadows.

Columns 3 and 4 of Fig.~\ref{fig:ablation_map} show the saliency maps of our method and those without GVM. When general visual features are introduced, the model can not only predict more complete salient objects, but also capture difficult-to-discern boundary parts, such as concave regions of the salient objects or protrude-out thin legs. 
In addition, we can see that our method generates more refined results when these low resolution general visual features are utilized, which suggests that our framework can effectively combine TRE with rich details and general visual features with abundant semantic information to produce excellent results.
%\textbf{Baseline}

\begin{table}
	%\small
	\centering
	\caption{Performance of GAS loss based on RGB features and general vision representations on DUTS-TE and PASCAL-S. ``GSA'' denotes the GSA loss, ``GVM'' denotes general visual module, ``GVR'' denotes general vision representations, and ``/'' indicates that GSA loss is not introduced.}
	\setlength{\tabcolsep}{1.5mm}{
		\begin{tabular}{l|c|ccc|ccc}
			%\multirow{2}{*}{Baseline}& \multirow{2}{*}{O}& \multirow{2}{*}{D} & \multirow{2}{*}{E} & \multicolumn{3}{|c|}  \\
			% &  &&  \\
			\toprule
			\multirow{2}*{Model}  &\multirow{2}*{GSA}  & \multicolumn{3}{c|} {DUTS-TE}& \multicolumn{3}{c} {PASCAL-S}   \\
			&&$F_{avg} $ &$MAE $ &$E_m $ &$F_{avg} $ &$MAE $ &$E_m $          \\
			%Baseline&SSL & Loss  &$F_{avg} \uparrow$ &$M \downarrow$ &$E_m \uparrow$ \\
			\midrule
			Baseline		  & /       &                 .823    & .049  &.890     &.818    & .077  &.846\\
			Baseline		  & RGB     &                 .826    & .048  &.890     &.821    & .075  &.843\\
			Baseline		  & GVR	    &                 .831    & .046  &.893     &.826    & .073  &.852\\
			Baseline + GVM    & / 	    &                 .838    & .039  &.906     &.835    & .075  &.873\\
			Baseline + GVM    & RGB     &                 .853    & .039  &.903     &.841    & .074  &.860\\
			Baseline + GVM    & GVR     &                 .861    & .037  &.908     &.859    & .071  &.877\\
			
			\bottomrule
	\end{tabular}}
	\label{table:color}
\end{table}
\subsubsection{Effectiveness of GSA Loss}
\begin{figure}[t]
	\centering
	\includegraphics[width=0.45\textwidth]{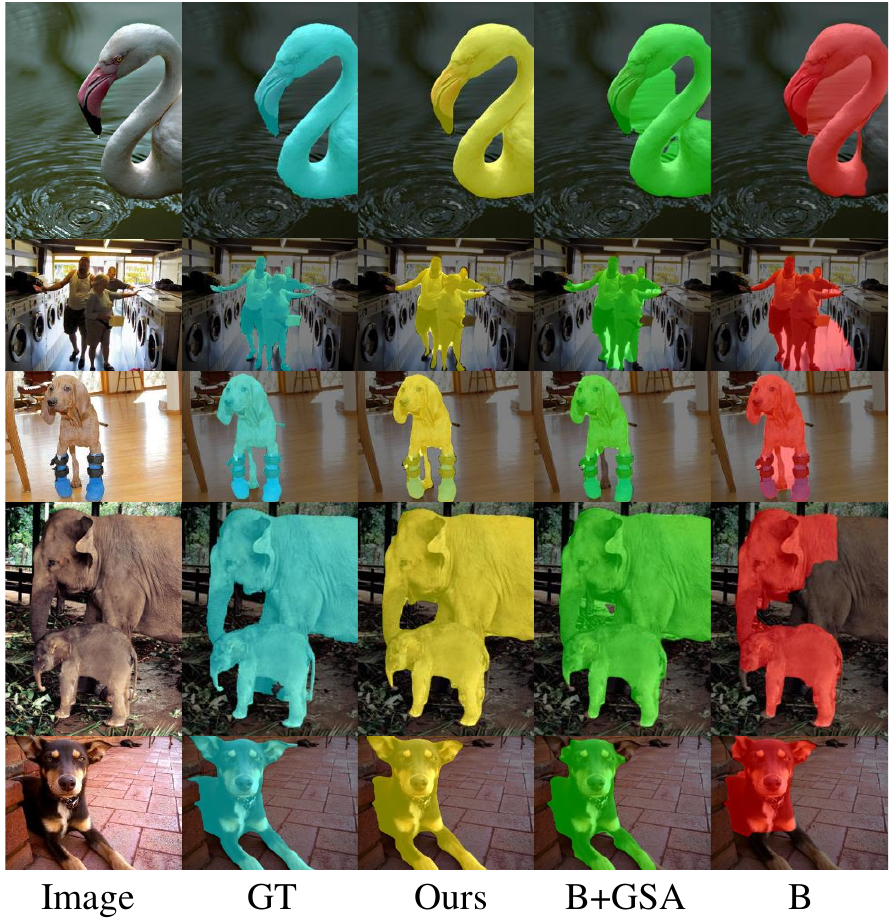} 
	\caption{Qualitative evaluation of general visual features and gsa loss. ``B'' denotes the baseline, ``B+GSA'' denotes the combination of the baseline and GSA loss, and ``Ours'' denotes the combination of the baseline, the GSA loss, and the GVM.}
	\label{fig:ablation_map}
\end{figure}

We evaluate the influence of our GSA Loss in Table~\ref{table:ablation_loss}.
It can be seen that when based on baseline model, using GSA loss can yield stable gain on all metrics, which shows that our loss can bring robust benefit to SOD.
Based on our model with fusing general visual features, using GAS loss can further improve the performance. This phenomenon explains general visual features rich in contextual semantic information, while our loss can further guide the model to exploit this information to focus on global affinities.
Columns 4 and 5 of Fig.~\ref{fig:ablation_map} show the saliency maps of the baseline with and without GSA loss.
Apparently, the predicted salient objects are more complete under the supervision of GSA loss.
Furthermore, we can observe that our method that introduces general visual features and GSA loss can predict more precise salient objects, which
means that our method can effectively play the respective roles of general visual representations and GSA loss.

%This method builds the similarity between nodes over the low-level color space, which may be effective to deal with the simple images, but is difficult for the complex scene of low foreground/background contrast.
In addition, to show the advantages of the GAS loss based on general visual representations, we compare our method with those based on low-level RGB color features, as shown in Table.~\ref{table:color}. We can find that the performance of the model is only slightly improved when the RGB-based GSA loss is introduced. However, adding the GSA loss based on general visual features can substantially improve the performance of the model. This result illustrates the effectiveness and rationality of the GSA loss based on general visual representations, which can address the issue that low-level features may be inadequate for complex scenes with low foreground-background contrast.

\begin{table}[t]
	%\small
	\centering
	\caption{Performance of point supervision and scribble supervision on DUTS-TE and PASCAL-S.}
	\setlength{\tabcolsep}{1.8mm}{
		\begin{tabular}{l|c|ccc|ccc}
			%\multirow{2}{*}{Baseline}& \multirow{2}{*}{O}& \multirow{2}{*}{D} & \multirow{2}{*}{E} & \multicolumn{3}{|c|}  \\
			% &  &&  \\
			\toprule
			\multirow{2}*{Model}  &\multirow{2}*{Sup.}  & \multicolumn{3}{c|} {DUTS-TE}& \multicolumn{3}{c} {PASCAL-S}   \\
			&&$F_{avg} $ &$MAE $ &$E_m $ &$F_{avg} $ &$MAE $ &$E_m $          \\
			%Baseline&SSL & Loss  &$F_{avg} \uparrow$ &$M \downarrow$ &$E_m \uparrow$ \\
			\midrule
			Baseline& Point   &                 .813    & .051  &.882     &.810    & .079  &.844\\
			Baseline& Scribble&                 .823    & .049  &.890     &.818    & .077  &.846\\
			Ours    & Point   &                 .850    & .042  &.901     &.830    & .077  &.855\\
			Ours    & Scribble&                 .861    & .037  &.908     &.859    & .071  &.877\\
			
			\bottomrule
	\end{tabular}}
	\label{table:point}
\end{table}

\begin{table}[t]
	%\small
	\centering
	\caption{Ablation analysis for different patch sizes of general visual features.}
	\setlength{\tabcolsep}{2.3mm}{
		\begin{tabular}{l|ccc|ccc}
			%\multirow{2}{*}{Baseline}& \multirow{2}{*}{O}& \multirow{2}{*}{D} & \multirow{2}{*}{E} & \multicolumn{3}{|c|}  \\
			% &  &&  \\
			\toprule
			\multirow{2}*{Patch Size}  & \multicolumn{3}{c|} {DUTS-TE}& \multicolumn{3}{c} {PASCAL-S}   \\
			&$F_{avg} $ &$MAE $ &$E_m $ &$F_{avg} $ &$MAE $ &$E_m $          \\
			%Baseline&SSL & Loss  &$F_{avg} \uparrow$ &$M \downarrow$ &$E_m \uparrow$ \\
			\midrule
			ViT-S/16&                  .847    & .040  &.910     &.832    & .076  &.865\\
			ViT-S/8 &                  .861    & .037  &.908     &.859    & .071  &.877\\
			
			\bottomrule
	\end{tabular}}
	\label{table:patch_size}
\end{table}

\subsubsection{Using Point Labels}
Scribble and point labels are quite similar in that they are both manually sparsely labeled and located inside salient objects and backgrounds. Therefore, to verify the reliability of our method further, we extend our method to the point-based SOD. In Table~\ref{table:point}, we compare our method with the baseline based on point annotations. The results show that our method based on point labels remarkably surpasses the point-based baseline, which further proves that our method is effective and robust.
Additionally, scribble-based methods are substantially superior to the point methods. The main reason is that scribble tags cover a wider range of salient objects and backgrounds, and can provide richer information to the model compared with point tags. It suggests that scribble tags may be a relatively cost-effective choice.

%\textbf{fignt!!!!!!!!!}

\subsubsection{Different Patch Sizes of General Visual Features}
Table~\ref{table:patch_size} shows the results of general visual features with different patch sizes. “16” and “8” denote patch sizes 16 and 8, respectively. The smaller the patch size, the higher the resolution of the general visual features. We can see that using general visual features with patch size of 8 performs better. When the resolution of the feature is increased, the model captures finer contextual connections and improves the details.

\subsubsection{Different Weights of GSA Loss}

\begin{table}
	%\small
	\centering
	\caption{Ablation analysis for different weight $\mu$ of the GAS loss.}
	\setlength{\tabcolsep}{2.3mm}{
		\begin{tabular}{c|ccc|ccc}
			%\multirow{2}{*}{Baseline}& \multirow{2}{*}{O}& \multirow{2}{*}{D} & \multirow{2}{*}{E} & \multicolumn{3}{|c|}  \\
			% &  &&  \\
			\toprule
			\multirow{2}*{$\mu$}  & \multicolumn{3}{c|} {DUTS-TE}& \multicolumn{3}{c} {PASCAL-S}   \\
			&$F_{avg} $ &$MAE $ &$E_m $ &$F_{avg} $ &$MAE $ &$E_m $          \\
			%Baseline&SSL & Loss  &$F_{avg} \uparrow$ &$M  \downarrow$ &$E_m \uparrow$ \\
			\midrule
			0.01&                  .852    & .039  &.902     &.848    & .074  &.872\\
			0.05&                  .859    & .038  &.909     &.856    & .072  &.879\\
			0.15&                  .861    & .037  &.908     &.859    & .071  &.877\\
			0.25&                  .862    & .037  &.908     &.857    & .071  &.875\\
			0.50&                  .842    & .043  &.894     &.821    & .085  &.845\\
			
			\bottomrule
	\end{tabular}}
	\label{table:weight}
\end{table}
%\textbf{Basic window size k is randomly chosen from a fixed numerical range to adjust the width of every synthetic concave region.} %有什么作用
$\mu$ is applied to adjust the weight of the GSA loss. $\mu$ is set to 0.01, 0.05, 0.15, 0.25, and 0.50,  and the experiments are conducted on DUTS-TE and PASCAL-S. Table~\ref{table:weight} shows when $\mu$ is given as 0.05, 0.15, and 0.25, the model achieves great performance. Finally, 0.15 is selected as the weight of the GSA loss because it performs slightly better among them.
%We finally select 0.3 as the weight of \textbf{l}  performs slightly better among them, 

\begin{figure}[h]
	\centering
	\includegraphics[width=0.45\textwidth]{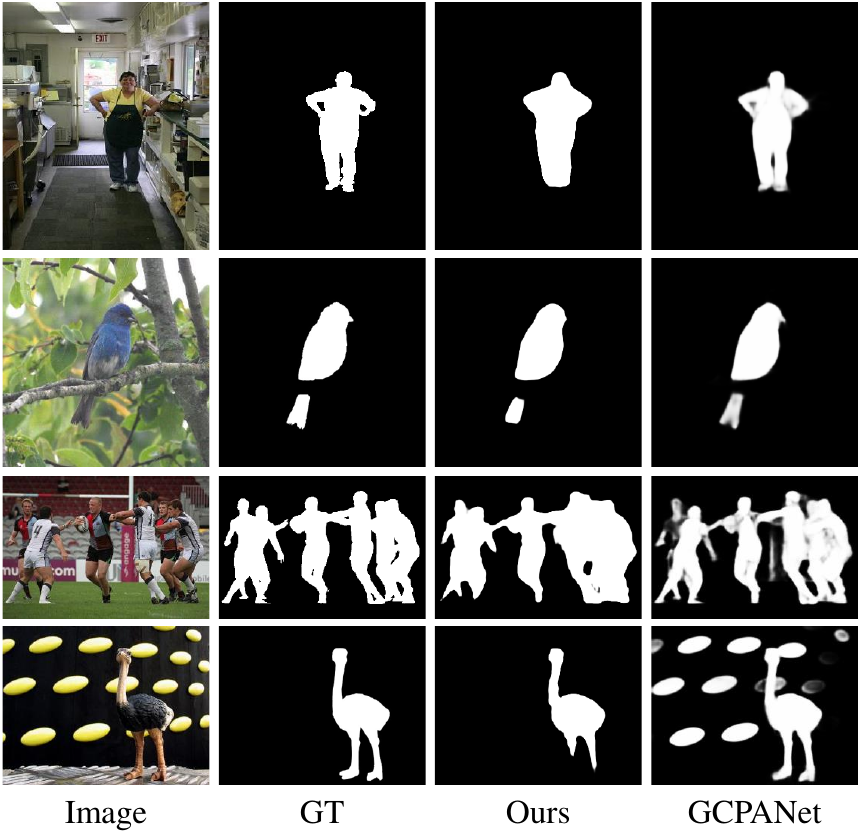} 
	\caption{Visual examples of the limitation of our method. Compared with the pixel-wise supervised method GCPANet, the edge parts of our predictions are relatively coarse.}
	\label{fig:limitation}
\end{figure}

\section{Limitations}
\label{sec::limitation}
Our proposed method mainly enhances the model's ability to distinguish salient objects by combining general visual representations and introducing the GSA loss, rather than optimizing edge details.
Thus, compared with pixel-wise supervised methods, there is still room for improvement in our method's ability to accurately capture object boundaries.
%Rows 1 and 2 of Fig.~\ref{fig:limitation} show the limitation of our method.
The first and second rows in Figure~\ref{fig:limitation} demonstrate the limitation of our method.
Our model is constructed based on the weakly supervised method SCWSSOD, which uses fully supervised model GCPANet as the basic network structure. Therefore, we directly compare our method with GCPANet to demonstrate the limitation of our method. 
It can be observed that compared with GCPANet trained on pixel-wise annotations, our method has difficulty in distinguishing precise object contours, such as the gaps between human legs and bird beaks.
Rows 3 and 4 provide an intuitive demonstration of both the strengths and limitations of our method. While our method excels in accurately detecting complete salient objects, it falls short in effectively capturing fine details at the edges. This is because although our method introduces general visual representations to help the model strengthen the connection between regions, the resolution of these features is relatively low. As a result, the model cannot significantly improve in boundary details.

\section{Conclusion}
\label{sec::conclusion}

%Previous models directly implement the SOD task only based on small-scale SOD training data, but due to the limited information provided by the scribble tags and the small-scale SOD training data, it is extremely difficult for the model to understand the image and further achieve a superior SOD task. 

%Inspired by human implementation of scribble-based SOD, this paper presents a simple yet effective framework guided by general visual representations, which aims to achieve a superior 

This work proposes a simple yet effective framework guided by general visual representations that simulate the general visual knowledge of humans for scribble-based SOD, which aims to address the difficulty that previous models cannot capture the accurate structure of salient objects only based on weak SOD training data without the support of general knowledge like humans. 
Besides, a GSA loss is further proposed by maintaining the high semantic similarities within the salient objects and within the backgrounds, and effectively guides the model to perceive global consistency. Without introducing any additional label, our method has made great progress, substantially surpassing previous weakly supervised SOD models, and is comparable with the state-of-the-art fully supervised method.
For future work, we will design an iterative refining framework to improve our method on edge details. Additionally, we will develop a general representation model for medical images and integrate it with our method to tackle the challenges in medical image segmentation.

%Previous models directly implement the SOD task only based on small-scale SOD training data without the support of rich experience and knowledge like humans, so it is difficult for them to break through the bottleneck.
%To , we propose a simple yet effective framework guided by general visual representations that simulate the general visual cognition of humans for scribble-based SOD.

%We propose a simple yet effective framework guided by general visual representations that simulate the general visual cognition of humans for scribble-based SOD.

%\section*{Acknowledgment}
%This work was partially supported by the National Key Research and Development Program of China (2020YFB1707700), the National Natural Science Foundation of China (62176235, 62036009, 61871350, U1909203), and Zhejiang Provincial Natural Science Foundation of China (LY21F020026).

%\ifCLASSOPTIONcaptionsoff
%\newpage
%\fi

{
	\bibliographystyle{IEEEtran}
	\bibliography{egbib}
}

\begin{IEEEbiography}
	[{\includegraphics[width=1in,height=1.25in,clip,keepaspectratio]{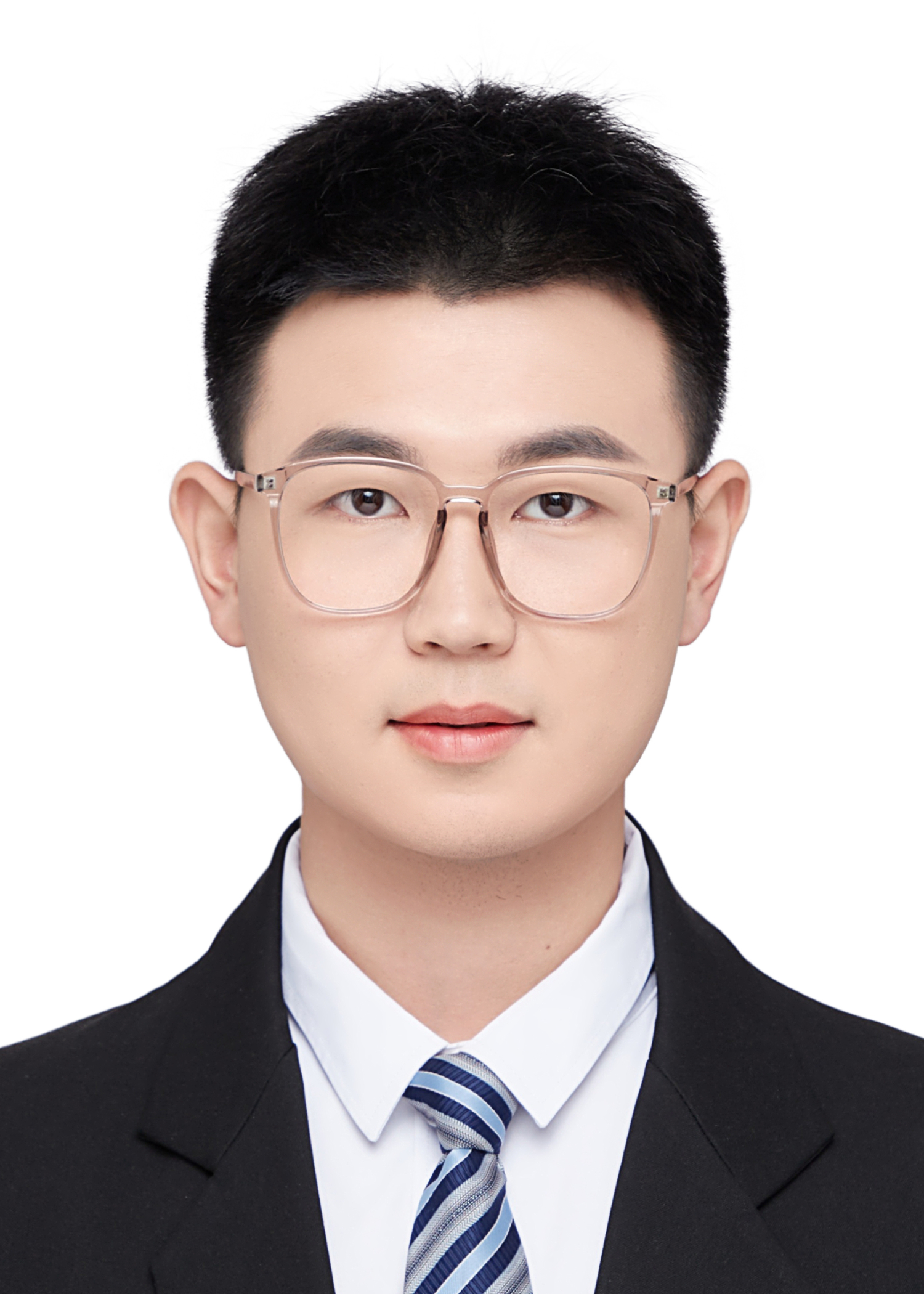}}]
	{Binwei Xu}
	received the B.Sc. in automation from Zhejiang University of Technology in 2018. He is currently pursuing the Ph.D. degree with the College of Computer Science, Zhejiang University of Technology. His area of interest lies in machine learning and computer vision.
\end{IEEEbiography}

\begin{IEEEbiography}
	[{\includegraphics[width=1in,height=1.25in,clip,keepaspectratio]{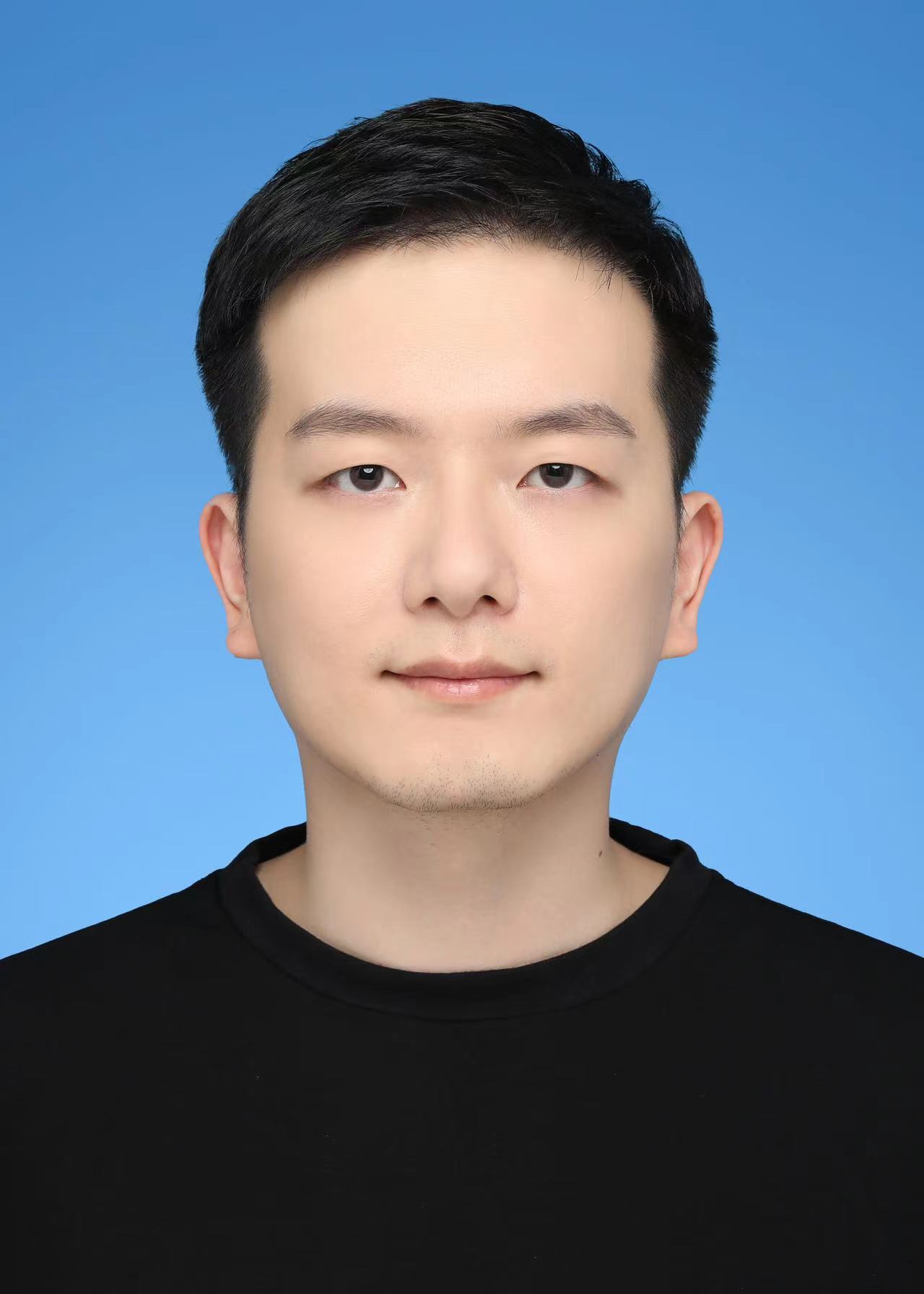}}]
	{Haoran Liang}
	received the B.Eng. degree in computer science from Zhejiang University of Technology in 2011, and the Ph.D. degree in control science and engineering from Zhejiang University of Technology in 2017. 
	He is currently a Lecture with the College of Computer Science, Zhejiang University of Technology, Hangzhou, China. His research interests include image/video saliency prediction, salient object detection, video summarization, image captioning and data visualization.
\end{IEEEbiography}

\begin{IEEEbiography}
	[{\includegraphics[width=1in,height=1.25in,clip,keepaspectratio]{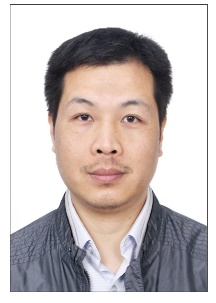}}]
	{Weihua Gong}
	received the Ph.D. degree in software engineering from HuaZhong University of Science and Technology in 2006, He is currently an associate professor, M.S. supervisor in the college of computer science, Zhejiang University of Technology, Hangzhou, China. His research interests include social network, machine learning.
\end{IEEEbiography}

\begin{IEEEbiography}
	[{\includegraphics[width=1in,height=1.25in,clip,keepaspectratio]{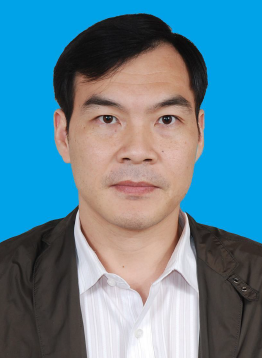}}]
	{Ronghua Liang}
	received the B.Sc. degree
	from Hangdian University, Hangzhou, China, in
	1996, and the Ph.D. degree in computer science from
	Zhejiang University, Hangzhou, China, in 2003.
	He worked as a Research Fellow with the University
	of Bedfordshire, Bedfordshire, U.K., from
	April 2004 to July 2005, and as a Visiting Scholar at
	the University of California, Davis, CA, USA, from
	March 2010 to March 2011. He is currently a Professor
	of computer science and the Dean
	of College of Computer Science with Zhejiang
	University of Technology. His research interests include computer vision, information
	visualization, and medical visualization.
	
\end{IEEEbiography}

% if you will not have a photo at all:
\begin{IEEEbiography}
	[{\includegraphics[width=1in,height=1.25in,clip,keepaspectratio]{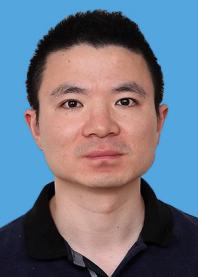}}]
	{Peng Chen}
	was born in Zhejiang Province, China, in 1981. He received his B.Sc. and Ph.D. degrees from Zhejiang University, Hangzhou, China, in 2003 and 2009 respectively. From 2015 to 2016, he was a visiting scholar with the University of California—Santa Barbara, Santa Barbara, CA, USA. He is currently a Professor with the Zhejiang University of Technology, Hangzhou, China. His research interests include computer vision, embedded systems design and pattern recognition.
\end{IEEEbiography}
\end{document}